\documentclass[letterpaper]{article} % DO NOT CHANGE THIS
\usepackage{aaai23}  % DO NOT CHANGE THIS
\usepackage{times}  % DO NOT CHANGE THIS
\usepackage{helvet}  % DO NOT CHANGE THIS
\usepackage{courier}  % DO NOT CHANGE THIS
\usepackage[hyphens]{url}  % DO NOT CHANGE THIS
\usepackage{graphicx} % DO NOT CHANGE THIS
\usepackage{natbib}  % DO NOT CHANGE THIS AND DO NOT ADD ANY OPTIONS TO IT
\usepackage{caption} % DO NOT CHANGE THIS AND DO NOT ADD ANY OPTIONS TO IT
\usepackage{algorithm}
\usepackage{algorithmic}

\usepackage{newfloat}
\usepackage{listings}
\DeclareCaptionStyle{ruled}{labelfont=normalfont,labelsep=colon,strut=off} % DO NOT CHANGE THIS
\floatstyle{ruled}
\newfloat{listing}{tb}{lst}{}
\floatname{listing}{Listing}
\usepackage{amsmath,amssymb}

\usepackage{bm}
\newcommand{\name}{DFMGAN}
\newcommand{\zobject}{\bm{z}_\mathrm{object}}
\newcommand{\zdefect}{\bm{z}_\mathrm{defect}}
\newcommand{\wobject}{\bm{w}_\mathrm{object}}
\newcommand{\wdefect}{\bm{w}_\mathrm{defect}}
\newcommand{\Fobject}{\bm{F}_\mathrm{object}}
\newcommand{\Fdefect}{\bm{F}_\mathrm{defect}}

\newcommand{\Dmatch}{D_\mathrm{match}}

\usepackage{xspace}
\makeatletter
\DeclareRobustCommand\onedot{\futurelet\@let@token\@onedot}
\def\@onedot{\ifx\@let@token.\else.\null\fi\xspace}

\def\eg{\emph{e.g}\onedot} 
\def\ie{\emph{i.e}\onedot}

\def\wrt{w.r.t\onedot}

\makeatother

\usepackage[capitalize]{cleveref}
\Crefname{section}{Section}{Sections}
\Crefname{table}{Table}{Tables}
\Crefname{figure}{Figure}{Figures}

\title{Few-Shot Defect Image Generation via Defect-Aware Feature Manipulation}
\author{
    Yuxuan Duan, Yan Hong, Li Niu$^{*}$, Liqing Zhang\thanks{Corresponding authors.}\\
}
\affiliations{
    MoE Key Lab of Artificial Intelligence\\
    Shanghai Jiao Tong University\\
    sjtudyx2016@sjtu.edu.cn, yanhong.sjtu@gmail.com, ustcnewly@sjtu.edu.cn, zhang-lq@cs.sjtu.edu.cn\\
}

\begin{document}

\maketitle

\begin{abstract}
The performances of defect inspection have been severely hindered by insufficient defect images in industries, which can be alleviated by generating more samples as data augmentation. We propose the first defect image generation method in the challenging few-shot cases. Given just a handful of defect images and relatively more defect-free ones, our goal is to augment the dataset with new defect images. Our method consists of two training stages. First, we train a data-efficient StyleGAN2 on defect-free images as the backbone. Second, we attach defect-aware residual blocks to the backbone, which learn to produce reasonable defect masks and accordingly manipulate the features within the masked regions by training the added modules on limited defect images. Extensive experiments on MVTec AD dataset not only validate the effectiveness of our method in generating realistic and diverse defect images, but also manifest the benefits it brings to downstream defect inspection tasks. Codes are available at \url{https://github.com/Ldhlwh/DFMGAN}.
\end{abstract}

%%%%%%%%%%%%%%%%%%%%
% INTRODUCTION
%%%%%%%%%%%%%%%%%%%%

\section{Introduction}
\label{sec:intro}

Defect inspection, whose typical tasks include defect detection, classification, and localization, plays an important role in industrial manufacture. So far, many research efforts have been paid to design automated defect inspection systems to ensure the qualification rate without manual participation \cite{review}. However, it is challenging to adequately obtain diverse defect images due to the scarcity of real defect images in production lines and the high collection cost, also known as the \emph{data insufficiency} issue. Therefore, nowadays deep learning-based defect inspection methods \cite{anogan,uninformed,cutpaste} usually adopt an unsupervised paradigm, that is, training one-class classifiers with defect-free data only. Without the supervision of defect images, those models cannot distinguish different defect categories and thus inapplicable to certain tasks such as defect classification.

\begin{figure}[t]
  \centering
  \includegraphics[width=\columnwidth]{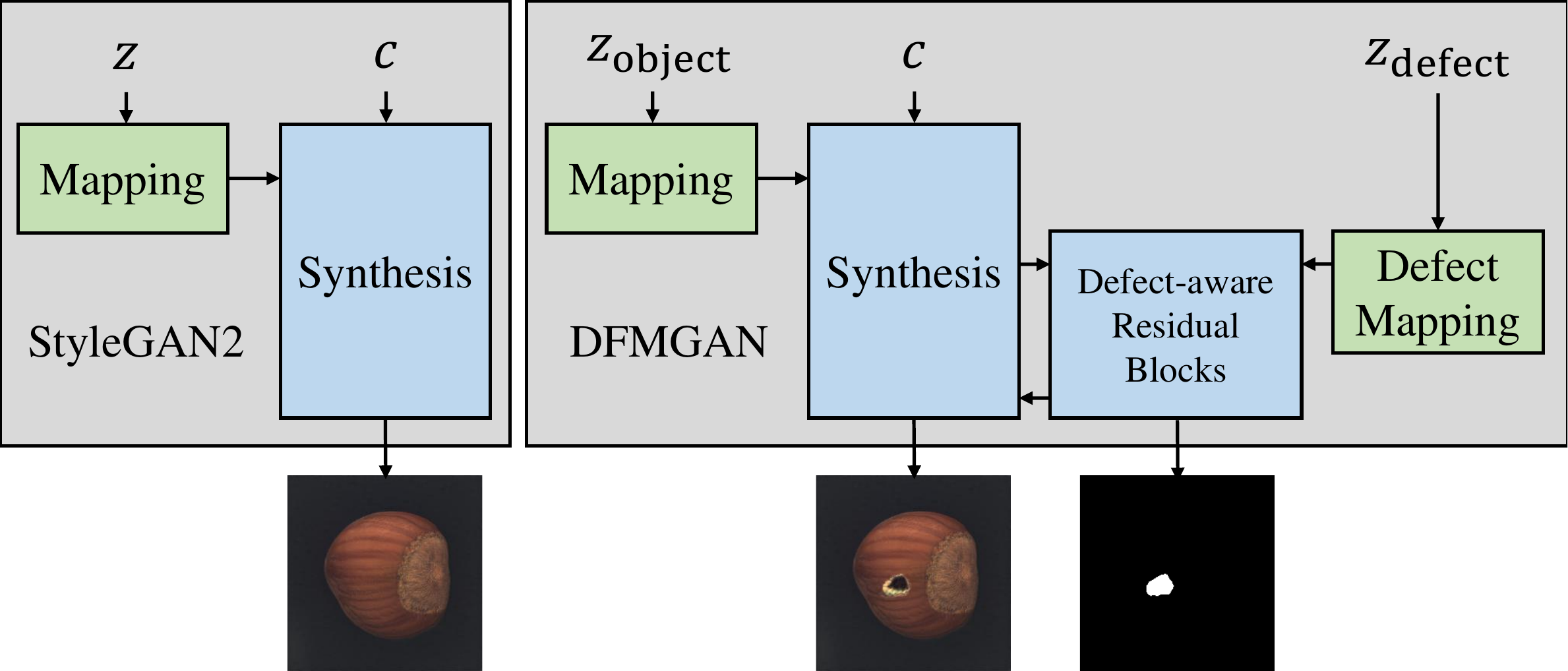}
   \caption{An overview of our \name{} and its two-stage training strategy. Left: First, a StyleGAN2 is pretrained on defect-free data. Right: Then, defect-aware residual blocks are attached to the backbone to produce defect masks and manipulate the features within defect regions.}
   \label{fig:intro}
\end{figure}

Aiming to solve the data insufficiency problem, an intuitive idea is to generate more defect images. Previous methods try to render simple yet fake defect images by manually adding artifacts \cite{cutout}, cutting/pasting patches of defect-free images (seen as defects) \cite{cutpaste}, or copying the defect region from one image to another \cite{nbr}. Nevertheless, defect images generated by these methods are far from being realistic and diverse. On the other hand, though Generative Adversarial Network (GAN) \cite{gan} and its variants are widely used in many image generation tasks, they are scarcely used for defect image generation because GANs are susceptible to data shortage. Previously, quite few GAN-based defect image generation works \cite{sdgan,defectgan} are designed. They either rely on hundreds or thousands of defect images and even more defect-free ones, or merely focus on a single category of texture (\eg, marble, metal, concrete). However, in real industrial manufacture, usually only a few defect images are available because of the rarity of real defect images in production lines and the difficulty in collection. 
%The defect-free images may not be affluent either. 
Moreover, comparing with textures, objects (\eg, nut, medicine, gadget) have richer structural information and fewer regular patterns, which further escalates the difficulty in defect image generation for objects.

To deal with such cases, we propose a novel Defect-aware Feature Manipulation GAN (\name{}) to generate realistic and diverse defect images using limited defect images. \name{} is inspired by few-shot image generation methods \cite{minegan,leverage,fsgan} which adapt pretrained models learned on large domains to smaller domains. However, these methods focus on transferring whole images without particular design on specific regions (\eg, defect areas in defect images). Based on the fact that a defective object has completely defect-free appearance except the defect regions, an intuitive idea could be adaptively adding defects to the generated defect-free images. In this work, with a backbone generator trained on hundreds of defect-free object images\footnote{For simplicity, we collectively call object and texture as \emph{object} when describing our model.}, we propose defect-aware residual blocks to produce plausible defect regions and manipulate the features within these regions, to render diverse and realistic defect images. An overview of the training process is shown in \cref{fig:intro}.
Extensive experiments on \emph{MVTec Anomaly Detection} (MVTec AD) \cite{mvtecad} prove that \name{} can not only generate various defect images with high fidelity, but also enhance the performance of defect inspection tasks as non-traditional augmentation.

Our contributions can be summarized as follows: 
(1) we make the first attempt at the challenging few-shot defect image generation task using a modern dataset MVTec AD with multi-class objects/textures and defects; 
(2) we provide a new idea of transferring critical regions rather than whole images which may inspire future works in many few-shot image generation applications; 
(3) we propose a novel model \name{} to generate realistic and diverse defect images associated with defect masks, via feature manipulation using defect-aware residual blocks; 
(4) experiments on MVTec AD dataset validate the effectiveness of \name{} on defect image generation and the benefits it brings to the downstream defect inspection tasks.

%%%%%%%%%%%%%%%%%%%%
% RELATED-WORK
%%%%%%%%%%%%%%%%%%%%

\section{Related Work}
\label{sec:related-work}

\paragraph{Defect Inspection}

Due to the data insufficiency issue, defect inspection methods cannot adopt a fully supervised paradigm. 
With the reconstruction and comparison strategy, AnoVAEGAN \cite{vaead} and AnoGAN \cite{anogan} utilized autoencoders \cite{dlbook} and GANs respectively. 
Besides these generative methods, \citet{cutpaste} used Grad-CAM \cite{gradcam} to show defect regions when identifying pseudo-defect images constructed from defect-free ones. \citet{uninformed} trained student networks imitating the output of a teacher network on defect-free data, and inferred a defect when obvious distinction between the students and the teacher occurs.

\paragraph{Image Generation on Limited Data}

Since proposed, Generative Adversarial Network (GAN) and its variants \cite{gan,cyclegan,starganv2,stylegan2} are renowned for the enormous data required to ensure the quality and diversity of generated images. There are some works focusing on training data-efficient GANs on small datasets. For instance, \citet{diffaug} proposed differentiable augmentation as a plugin to StyleGAN2 \cite{stylegan2}. Nevertheless, these works still generally required at least hundreds of images, leaving directly training on just several or tens of images unsolved. Some other works tried to transfer the model pretrained on larger datasets to boost its performance on small datasets. For example, \citet{batchstat,leverage,fsgan} eased the transfer process by limiting the number of trained parameters. \citet{minegan} explored the transferable latent space regions of the generator. \citet{cross} preserved a one-to-one correspondence with cross-domain consistency loss.
These methods transferred the distribution of the whole images, while we suppose transferring only specific regions (\ie defect regions) may be beneficial to our task.

\paragraph{Defect Image Generation}

The rarity of defect samples has motivated research efforts on defect image generation as data augmentation for defect inspection applications. \citet{cutout} added random cutouts on normal images as artificial defects. \citet{cutpaste} copied a patch from a defect-free sample and pasted it to another location, rendering a pseudo-defect. \citet{nbr} cropped the defect regions of a defect image and pasted it to another defect-free one. Among these non-generative methods, the first two utilized defect-free images only, whose generated samples are not category-specific thus not applicable to inspection tasks such as defect classification. Crop\&Paste \cite{nbr} was only able to yield limited number of defect samples depending on the size of datasets, and actually it could not generate new defects, but moved in-dataset defects onto different objects. Also, traditional data augmentation can be hardly used on defect images because few transformations (\eg, flipping, rotation) keep intact defects without affecting color, pattern, position and other characteristics.

To the best of our knowledge, only two previous works \cite{sdgan,defectgan} designed generative augmenting methods. \citet{sdgan} proposed SDGAN, translating defect-free and defect images interchangeably through two generators. Similarly, \citet{defectgan} simulated defacement and restoration processes by adding and removing defect foregrounds using Defect-GAN. However, these two works had certain limitations:
(1) \textbf{Large texture datasets}: They had access to hundreds or thousands of defect samples of a single category, which are not always accessible. Also, the datasets they used are on highly specific textures (cylinder surfaces of commutators or concrete surfaces), which had much less structural information than objects (\eg, hazelnuts as we use for the experiments).
(2) \textbf{Merely generate defects}: Both works needed defect-free samples as their input while rendering defects via image-to-image paradigm. This strategy limited the diversity of the object/texture backgrounds, especially in cases that defect-free images were not abundant either.
(3) \textbf{Lack randomness}: SDGAN did not involve randomly sampled codes or noises as GANs usually do, which further limited the diversity.
(4) \textbf{No masks}: Neither of these works produced defect masks with clear boundaries, restricting their usage in certain inspection tasks (\eg, defect localization) requiring ground-truth defect masks.

To tackle the aforementioned limitations of previous works, in the following sections, we will introduce \name{}, which is the first few-shot defect image generation method capable of rendering realistic images with high diversity on both objects and defects.

%%%%%%%%%%%%%%%%%%%%
% METHOD
%%%%%%%%%%%%%%%%%%%%

\begin{figure*}[t]
  \centering
  \includegraphics[width=0.9\textwidth]{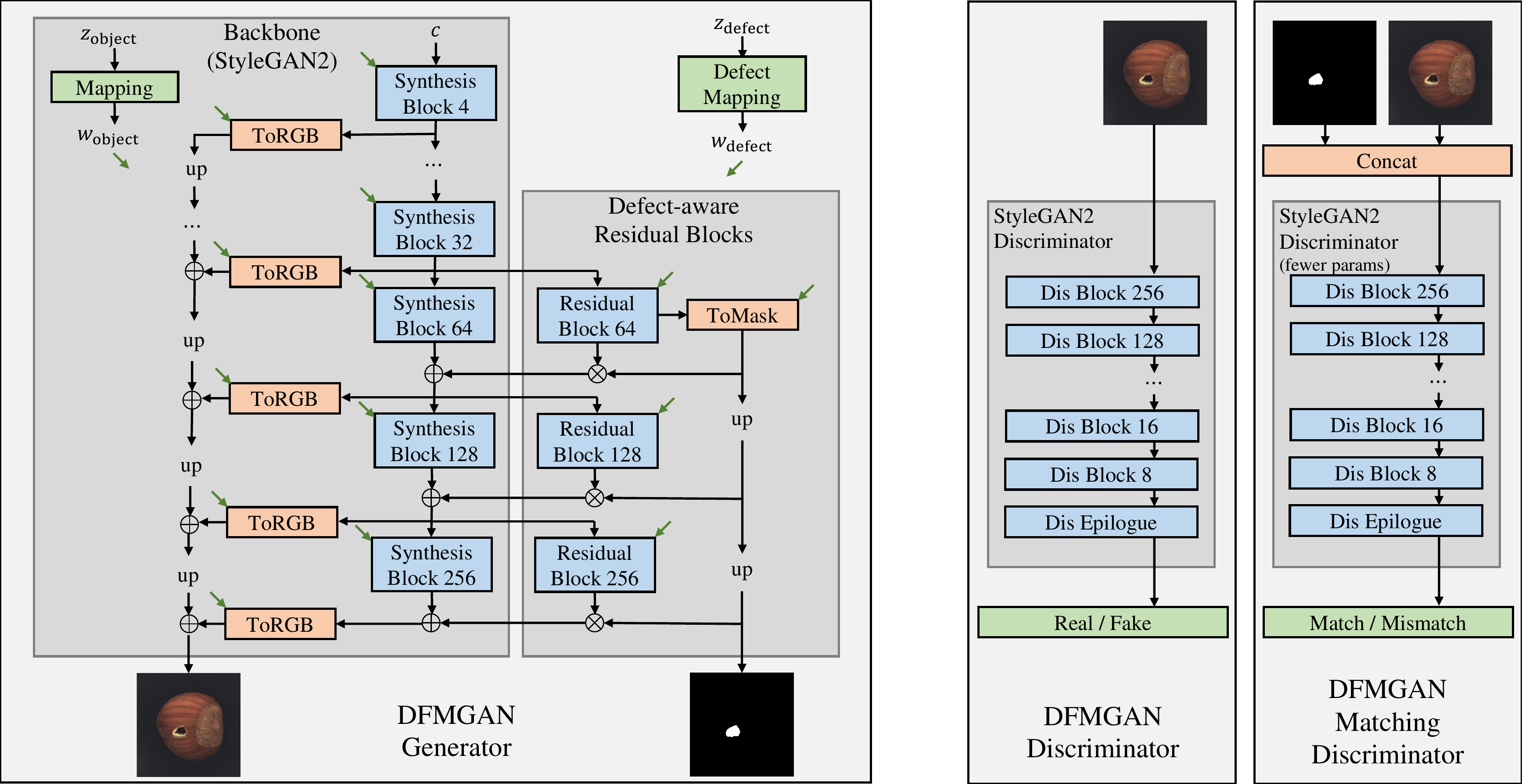}
   \caption{The architecture of \name{}. Left: The generator mainly consists of the backbone and the defect-aware residual blocks. The backbone adopts the original structure of StyleGAN2 with \emph{ToRGB} modules accumulating RGB appearances in the \emph{skip} manner. The defect-aware residual blocks manipulate the features starting from resolution $64 \times 64$, with masks from \emph{ToMask} module controlling the manipulating areas. Both parts have weights modulated by their corresponding mapping networks shown as green arrows. Right: The two discriminators, respectively in charge of judging the realism of images and whether the defects in images match the masks. We generally remain the original structure of StyleGAN2 discriminator.}
   \label{fig:dfmgan}
\end{figure*}

\section{Method}
\label{sec:method}

As shown in \cref{fig:dfmgan}, \name{} adopts a two-stage training strategy. First, we train a data-efficient StyleGAN2 as the backbone on hundreds of defect-free images, which maps a random code $\zobject$ to an image without defect (\cref{sec:backbone}). Second, we attach defect-aware residual blocks along with defect mapping network to the backbone, and train these added modules on a few defect images. The entire generator maps $\zobject$ and a defect code $\zdefect$ to defect images with controllable defect regions (\cref{sec:residual}). 

\subsection{Pretraining on Defect-free Images}
\label{sec:backbone}
In the first training stage, we aim to train a generator as the backbone of \name{} to produce diverse defect-free images by randomly sampling object codes $\zobject$. Considering the superiority of generation ability of StyleGANs, we adopt StyleGAN2 with Adaptive Differentiable Augmentation \cite{stylegan2-ada} as the backbone, which consists of a mapping network and a synthesis network. The synthesis network, taking a learned constant feature map $c$, is composed of convolutional synthesis blocks with the \emph{skip} architecture accumulating RGB appearances through \emph{ToRGB} modules to the final generated images. The mapping network takes a random $\zobject$ and maps it to $\wobject$ which modulates the convolution weights of the synthesis network (green arrows in \cref{fig:dfmgan}), importing variations to the generated images. Besides the generator, a discriminator is also trained to provide supervision. Refer to \citet{stylegan2} for detailed designs of StyleGAN2. 
After this stage, the backbone generator encodes rich object features in its network. In the next stage, we will attach defect-aware residual blocks, which can adapt the model from defect-free images to defect ones.

\subsection{Transferring to Defect Images}
\label{sec:residual}
Considering the fact that a defect image is composed of defect regions and defect-free regions, we conjecture that by properly manipulating the potential defect regions of the object feature maps from the backbone, the whole model can be extended to produce defect images while maintaining the generation ability of defect-free images. Motivated by this idea, in the second training stage, we propose the defect-aware residual blocks attached to the backbone, rendering plausible defect masks delimiting defect regions and the corresponding defect features. The masks and the defect features are then used to manipulate the object feature maps to add defects to defect-free images. To ensure the fidelity and variation of the defects, we further employ an extra defect matching discriminator and a modified mode seeking loss respectively during this stage.

\paragraph{Defect-aware Residual Blocks}
Our proposed defect-aware residual blocks share similar structure with the synthesis blocks of the backbone.
At resolution 64 where the first residual block is attached, the synthesis block $S$ and the residual block $R$ both take the feature $\bm{F}^{32}$ from the last synthesis block at resolution 32 and output object feature map $\Fobject^{64} = S(\bm{F}^{32})$ and defect residual feature map $\Fdefect^{64} = R(\bm{F}^{32})$ respectively, where $\Fobject^{64}, \Fdefect^{64} \in \mathbb{R}^{N \times 64 \times 64}$ and $N$ is the number of channels. Then the extra \emph{ToMask} module, like its counterpart \emph{ToRGB} modules of the backbone, determines the defect region of the current image by generating a mask $\bm{M} = \emph{ToMask}(\Fdefect^{64}) \in \mathbb{R}^{64 \times 64}$.
Only the residual features corresponding to the defect pixels (non-negative values on the mask $\bm{M}$) are added to the object feature map, leading to the manipulated feature map $\bm{F}^{64}$:
\begin{equation}
  \bm{F}^{64}(i,j) = \left\{
  \begin{aligned}
    &\Fobject^{64}(i,j)+\Fdefect^{64}(i,j), &  & \bm{M}(i,j) \ge 0, \\
    &\Fobject^{64}(i,j), & & \bm{M}(i,j) < 0,
  \end{aligned}\right.
  \label{eq:mask}
\end{equation}
where $(i, j)$ represents any pixel on the feature map or the mask.
In this way, the residual blocks only manipulate the object features within the defect regions, and those in non-defect areas remain unchanged. The manipulated feature map $\bm{F}^{64}$ then takes the place of $\Fobject^{64}$ to be the input of the following blocks.
Later at resolution 128 and 256, the mask $\bm{M}$ is upsampled to the corresponding resolution to control the defect residual features, which further manipulate the object feature maps within the defect regions in a similar way to resolution 64. 
We leave the synthesis blocks at resolution 32 or lower untouched because the high-level layers (with lower resolution) of the networks decide the coarse structure of the images, while low-level layers generate detailed appearances including the defects \cite{leverage}. 
To ensure the diversity of the generated defect images, instead of being solely determined by the object feature map from the backbone, we introduce an extra defect mapping network to control the variation of defects. The defect mapping network takes in a randomly sampled defect code $\zdefect$ and outputs the modulation weights $\wdefect$, which is used to modulate the residual blocks (green arrows in \cref{fig:dfmgan}) similar to the backbone. The two mapping networks share the same network structure.

During the second training stage, \name{} fixes its backbone and trains the defect mapping network along with our proposed defect-aware residual blocks on defect images in order to generate more defect samples with high fidelity and diversity. We control the number of trainable parameters in this stage to 3.7M. Compared with the fixed backbone of 23.2M trainable parameters in the previous stage on defect-free images, it will be much easier to train on just a handful of defect images in the second stage. Another advantage of \name{} is that, by fixing the parameters of the backbone, it retains the ability of generating defect-free images as long as we cut off the defect-aware feature manipulation by ignoring the defect residual features from the residual blocks. 

\paragraph{Two Discriminators}
Due to the content similarity between the defect-free images and the defect images, we can easily transfer the pretrained discriminator from defect-free images to defect ones by finetuning. Yet, this discriminator can only provide supervision to the images, not the masks. 
To guarantee that the generated defect masks precisely delimit the defect regions of the images, we use an extra defect matching discriminator $\Dmatch$ to bridge the gap between real pairs of defect image and mask and generated pairs.
$\Dmatch$ has almost the same architecture with the original discriminator $D$, but we reduce the number of channels in each layer based on the intuition that judging whether a defect image matches a mask is easier than judging its realism.
With much fewer parameters (1.5M comparing with 24M of $D$), $\Dmatch$ is suitable for few-shot defect image generation. Pairs of image and mask are concatenated before being fed into $\Dmatch$. The output scores judge whether these defect images match their corresponding defect masks. $\Dmatch$ provides supervision by optimizing Wasserstein adversarial loss \cite{wgan} with R1 regularization as $D$ does in StyleGAN \cite{stylegan1}.
The two discriminators cooperate with each other in the process of defect image generation.

\paragraph{Mode Seeking Loss}
In our model, the generated defect images depend on the object features from the backbone and the defect features from the defect-aware residual blocks. This design matches the fact that the defect on an object depends on both the object itself and the external factors. 
However, preliminary experiments with \name{}, where we vary $\zdefect$ yet fix $\zobject$ (thus also fix the object features), have shown that the defects are almost merely determined by the object features, with hardly noticeable changes when using different $\zdefect$. In this way, it can be foreseen that similar objects will always be accompanied by resembling defects, which substantially harms the diversity.

To mitigate this problem, we employ a variant of the mode seeking loss \cite{modeseek} in the second training stage. With two random defect codes $\zdefect^1$ and $\zdefect^2$, the defect mapping network outputs two corresponding modulation weights $\wdefect^1$ and $\wdefect^2$. The whole model produces defect masks $\bm{M}^1$ and $\bm{M}^2$ respectively using $\wdefect^1$ and $\wdefect^2$ along with the same $\wobject$.
Then, \name{} minimizes the mode seeking loss
\begin{equation}
    L_\mathrm{ms} = \frac{\|\wdefect^1 - \wdefect^2\|_1}{\|\bm{M}^1 - \bm{M}^2\|_1}.
\end{equation}
In other words, when using different $\wdefect$, we hope that the difference between the defect masks is maximized. 

Practically, we use $\wdefect$ instead of $\zdefect$ because \citet{stylegan1} states that the latent space of $\bm{w}$ is less entangled and hence better to represent the input space of the generator than a fixed distribution of $\bm{z}$. Also, due to the unexpected artifacts on the defect appearances, we use the differences between masks instead of images. See the ablation study for details.

\paragraph{Objective}
Given the original loss function $L_\mathrm{StyleGAN}$ used by StyleGAN2 (including adversarial loss, path length regularization and R1 regularization, refer to \citet{stylegan1}), the loss function $L_\mathrm{match}$ of $\Dmatch$ and the mode seeking loss $L_\mathrm{ms}$, our \name{} alternatively optimizes the generator $G$ and the discriminators $D, \Dmatch$ according to the overall objective function
\begin{equation}
    \begin{aligned}
        L(G, D, \Dmatch) = & L_\mathrm{StyleGAN}(G, D) + \\ 
        & L_\mathrm{match}(G, \Dmatch) + \lambda L_\mathrm{ms}(G),
    \end{aligned}
\end{equation}
where setting $\lambda = 0.1$ generally renders good results.

%%%%%%%%%%%%%%%%%%%%
% EXPERIMENT
%%%%%%%%%%%%%%%%%%%%

\begin{table*}[t]
  \centering
  \begin{tabular}{lrr|rr|rr|rr}
    \hline
    Defect & \multicolumn{2}{c|}{Crack} & \multicolumn{2}{c|}{Cut} & \multicolumn{2}{c|}{Hole} & \multicolumn{2}{c}{Print} \\
    %\hline
    Method & KID$\downarrow$ & LPIPS$\uparrow$ & KID$\downarrow$ & LPIPS$\uparrow$ & KID$\downarrow$ & LPIPS$\uparrow$ & KID$\downarrow$ & LPIPS$\uparrow$ \\
    \hline
    Finetune & 41.64 & 0.1541 & 21.80 & 0.1192 & 30.54 & 0.1263 & 28.75 & 0.1526 \\
    DiffAug & 24.69 & 0.0570 & 19.84 & 0.0456 & 22.43 & 0.0466 & 39.03 & 0.0604 \\
    CDC & 206.14 & 0.0437 & 213.98 & 0.0390 & 271.72 & 0.0566 & 355.37 & 0.0500 \\
    \hline
    Crop\&Paste & - & 0.1894 & - & 0.2045 & - & 0.2108 & - & 0.2185 \\
    \hline
    SDGAN & 148.86 & 0.1607 & 161.16 & 0.1474 & 152.86 & 0.1689 & 176.09 & 0.1748 \\
    Defect-GAN & 30.98 & 0.1905 & 32.69 & 0.1734 & 36.30 & 0.2007 & 33.35 & 0.2007 \\
    \textbf{\name{}} & \textbf{19.73} & \textbf{0.2600} & \textbf{16.88} & \textbf{0.2073} & \textbf{20.78} & \textbf{0.2391} & \textbf{27.25} & \textbf{0.2649} \\
    \hline
  \end{tabular}
  \caption{The results of the few-shot defect image generation on the object category \emph{hazelnut}, where we report KID$\times 10^3$@5k and clustered LPIPS@1k for each setting. The three groups of methods are respectively generic few-shot image generation methods, non-generative and generative defect image generation methods. \name{} outperforms in all four defect categories \emph{crack}, \emph{cut}, \emph{hole}, \emph{print}. For other object/texture categories, see \emph{supplementary material}.}
  \label{tab:gen}
\end{table*}

\section{Experiment}
\label{sec:experiment}

To verify the effectiveness of \name{}, we conduct experiments on the dataset MVTec AD (\cref{sec:mvtecad}), including the defect image generation task (\cref{sec:exp-dig}) and the downstream defect classification task (\cref{sec:exp-dc}). See \emph{supplementary material} for implementation details and the ablation study validating several choices in designing our model.

\subsection{Dataset: MVTec AD}
\label{sec:mvtecad}

MVTec Anomaly Detection\footnote{\url{https://www.mvtec.com/company/research/datasets/mvtec-ad}, released under CC BY-NC-SA 4.0.} (MVTec AD) \cite{mvtecad} is an open dataset containing ten object categories and five texture categories commonly seen, with up to eight defect categories for each object/texture category. All the images are accompanied with pixel-level masks showing the defect regions. Although originally designed for defect localization, MVTec AD fits the experimental setting for few-shot defect image generation since most object/texture categories have 200--400 defect-free samples, and most defect categories have 10--25 defect images. In the first stage, the backbone of \name{} is trained on the defect-free images of an object/texture category in the training set. In the second stage, an individual \name{} is trained for each defect category associated with this object/texture category. All images are resized to a moderate resolution of $256 \times 256$.

In the main paper, we mainly focus on the object category \emph{hazelnut}, which is a highly challenging category in MVTec AD due to its naturally high variation and complex appearance compared to the other manufactured objects. It has four defect categories \emph{crack}, \emph{cut}, \emph{hole}, and \emph{print}.
We leave the results on the other categories to \emph{supplementary material}.

\subsection{Defect Image Generation}
\label{sec:exp-dig}

\paragraph{Metric}
Following \citet{stylegan2-ada}, we use Kernel Inception Distance (KID) \cite{kid} as our main metric. KID resembles the conventionally used metric Fr\'echet Inception Distance (FID) \cite{fid} in image generation tasks, yet is designed without bias and thus a more descriptive metric on small datasets. Similar to FID, KID evaluates both the reality and the diversity of the generated images where lower values indicate better performance. We report KID between 5,000 generated defect images and the defect images of the corresponding defect category in the dataset.

However, KID (as well as FID) is commonly observed to prefer reality to diversity.
Therefore, to supplement the experiment results with a standalone diversity metric, we also report a clustered version of Learned Perceptual Image Patch Similarity (LPIPS) \cite{lpips} used by recent few-shot image generation works. %\cite{cross}%
Suppose the dataset of a defect category contains $N$ images. First 1,000 generated images are grouped into $N$ clusters by finding the closest (lowest LPIPS) dataset image, then we compute the mean pairwise LPIPS within each cluster and finally compute the average of them. Such clustered LPIPS suits few-shot image generation tasks because overfitted models will receive nearly zero scores, and higher scores indicate better diversity.

\paragraph{Baseline}

We compare our \name{} with six other methods. \emph{Finetune} pretrains and finetunes both using the original StyleGAN2. \emph{DiffAug} \cite{diffaug} uses differentiable data augmentation to prevent overfitting when directly training on small datasets. \emph{CDC} \cite{cross} preserves the cross-domain correspondence among the source/target images using consistency loss. These three are generic few-shot image generation methods not specialized in generating defects. 
\emph{SDGAN} \cite{sdgan} and \emph{Defect-GAN} \cite{defectgan} are the only two generative methods for defect image generation tasks prior to our work, whose details are discussed in \cref{sec:related-work}.
Finally, though not a generative model, we include \emph{Crop\&Paste} \cite{nbr} as the representative of non-generative methods to make the experiments comprehensive.

\begin{figure*}[t]
  \centering
  \includegraphics[width=1.0\linewidth]{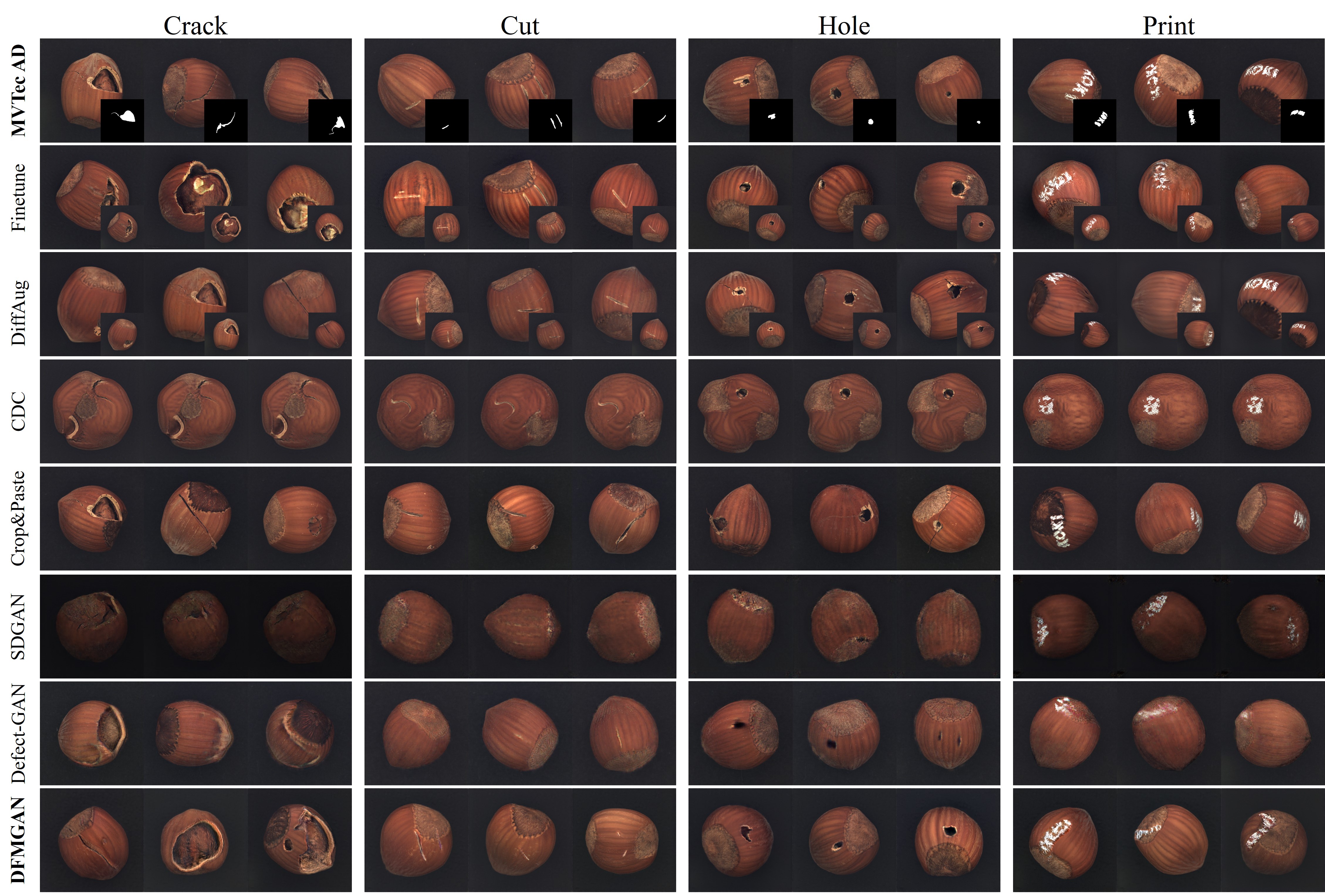}
   \caption{Examples of datasets (with defect masks) and generated defect images. Finetune and DiffAug are  prone to overfit the dataset (we show their closest images in the dataset), while CDC generates unreal images without much differences. Crop\&Paste cannot produce new defects (\eg, the first \emph{crack} image has the defect appearance of the first one in MVTec AD) and sometimes the defects go beyond the boundaries of the objects (\eg, the first \emph{hole} image). SDGAN and Defect-GAN fail to render realistic samples. \name{} has the most satisfying performance balancing quality and diversity. See \emph{supplementary material} for other categories.}
   \label{fig:gen}
\end{figure*}

\paragraph{Quantitative Result}

The KID and clustered LPIPS results of defect image generation are shown in \cref{tab:gen}. Note that since the produced images of Crop\&Paste have almost the same distribution \wrt appearance with the datasets, they always receive nearly zero KID scores which are omitted. For all the defect categories of hazelnut, our \name{} outperforms all the other methods on both KID and clustered LPIPS, showing its strong ability in rendering defect images with high quality and diversity despite of the severely insufficient data it is trained on.

\paragraph{Qualitative Result}

To provide a visual comparison among \name{} and the other methods, we show examples of the generated defect images in \cref{fig:gen}. Although the images yielded by Finetune and DiffAug have good quality, they are actually overfitting the training images, contributing marginal extra diversity. In contrast, CDC, SDGAN and Defect-GAN cannot produce realistic samples in these few-shot cases on the challenging object category. Crop\&Paste only borrows the defect appearances from the datasets. Finally, our \name{} achieves a good balance between reality and diversity, generating satisfying images.

\begin{figure}[t]
  \centering
  \includegraphics[width=0.8\columnwidth]{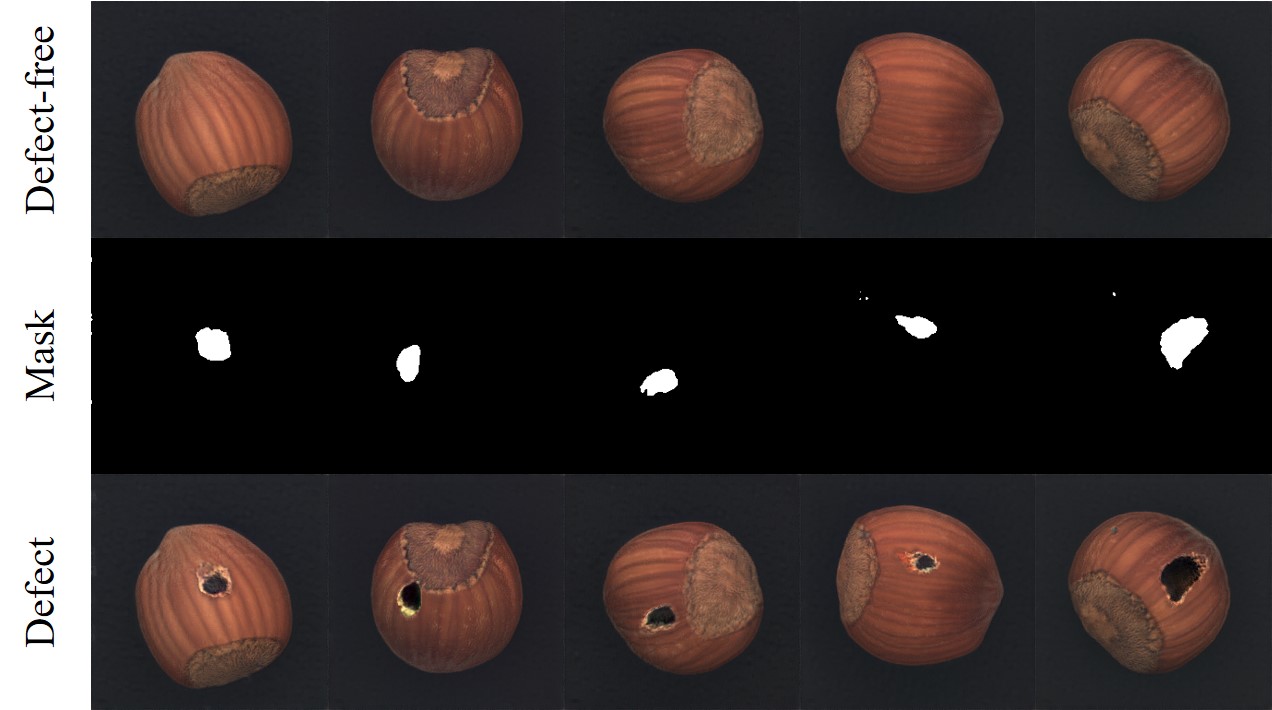}
   \caption{Examples of the triplets of generated defect-free image, mask and defect image. Our \name{} can render paired defect/defect-free images with difference only in the defect regions precisely delimited by the pixel-level masks.}
   \label{fig:cmp}
\end{figure}

For our method, we also show groups of generated defect-free image, defect image and its mask of defect category \emph{hole} in \cref{fig:cmp}, where the masks precisely show the defect regions. Also, by fixing the parameters of the backbone in the second training stage, \name{} is still able to generate defect-free samples if the model is forced to ignore the defect-aware residual features. Paired defect-free and defect images can be utilized in defect-related tasks such as defect restoration.

\paragraph{Discussion}

Comparing the few-shot image generation methods (Finetune, DiffAug, CDC) and the previous generative defect generation methods (SDGAN, Defect-GAN), we have found that the formers, without special designs for generating defects, are prone to overfit. Most images yielded by Finetune or DiffAug are roughly identical to one of the defect images from the dataset, thus generally achieving good KID scores but relatively low clustered LPIPS. On the other hand, CDC, originally tested on human face datasets, suffers from the close appearances of the defect-free images, hence it cannot render defect images with much variations either. Without much improvement to the diversity of the augmented dataset, these methods fail to provide helpful information for the downstream task in the next section.

On the contrary, SDGAN and Defect-GAN are able to ensure their diversity to a certain degree by generating defect images based on the relatively more defect-free data. However, we suppose that the critical flaw in the designs of these two methods is their image-to-image paradigm. They still have to guarantee the quality of the non-defect areas while rendering defects simultaneously, which is a harder task than focusing only on the defects explicitly delimited by the generated defect masks. Hence, the generated samples by SDGAN or Defect-GAN in \cref{fig:gen} expose quality deterioration to the objects. As \cref{tab:gen} shows, these two methods receive worse KID, and the generated images are far from being realistic to help the downstream tasks.

For the non-generative method Crop\&Paste, it can only render a finite number of defect images. For a dataset including $\bm{N}_g$ defect-free images and $\bm{N}_d$ defect images, Crop\&Paste is unable to generate more than $\bm{N} = \bm{N}_g \times \bm{N}_d$ samples. Therefore, though it achieves rather high clustered LPIPS scores in \cref{tab:gen} since $\bm{N} > 1000$ for the defect categories of hazelnut, it can be foreseen that the diversity will worsen when we require more than $\bm{N}$ images. Besides, as in \cref{fig:gen}, defects produced by Crop\&Paste sometimes (partially) locate outside the object because the generation process of this method violates the fact that defects are \emph{co-decided} by the objects and external factors.

We have designed our \name{} aiming to settle the aforementioned issues. We train our model on defect-free images in the first training stage to capture the distribution of the object category, which is beneficial to be transferred to the defect categories. In the second stage, our model is forced to learn to add realistic defects on the features of various defect-free samples learned in the first stage, preventing overfitting and enhancing the diversity of the generated defect images. In addition, \name{} can merely focus on the defect regions since the non-defect areas are determined by $\zobject$ and the fixed backbone, which keeps the overall reality of our generated defect samples. In conclusion, with our specially designed architecture, \name{} can handle the challenging few-shot defect image generation cases even on objects with complex structural information and high variation, and outperform previous methods by a large margin. 

\paragraph{Few-shot Generation}

To check the performance of \name{} in extreme cases with even fewer defect samples, we further challenge our model on 5-shot/1-shot defect image generation. We leave this part to \emph{supplementary material}.

\subsection{Data Augmentation for Defect Classification}
\label{sec:exp-dc}

\begin{table}[t]
  \centering
  \begin{tabular}{lrrr}
    \hline
    Method & P1 Acc$\uparrow$ & P2 Acc$\uparrow$ & P3 Acc$\uparrow$ \\
    \hline
    Finetune & 70.83 & 72.91 & 70.83 \\
    DiffAug & 64.58 & 62.50 & 68.75 \\
    CDC & 58.33 & 64.58 & 41.67 \\
    \hline
    Crop\&Paste & 66.67 & 52.08 & 58.33 \\
    \hline
    SDGAN & 56.25 & 31.25 & 43.75 \\
    Defect-GAN & 60.42 & 68.75 & 54.17 \\
    \textbf{\name{}} & \textbf{83.33} & \textbf{81.25} & \textbf{81.25} \\
    \hline
  \end{tabular}
  \caption{The results of the defect classification experiments for the object category \emph{hazelnut}. The training images generated by \name{} achieve the best accuracies on classifying unseen defect samples in all three partitions P1--3.}
  \label{tab:class}
\end{table}

Defect classification is a fundamental defect inspection task recognizing different types of defects of one object category. Since none of the baseline methods (except Crop\&Paste) renders clear masks showing the defect regions, comparison on mask-requiring tasks such as defect localization is impossible. Thus we choose to test \name{} on defect classification which does not require masks. Nevertheless, \name{} can serve as a baseline in other tasks for future works.

For these experiments, we randomly choose 1/3 of the dataset images from each defect category as the base sets, and the other 2/3 from each category are combined as the test set. As for the hazelnut category, each base set has five or six images, and the test set consists of 12 images from each defect category, 48 in total. We train the methods on the four base sets corresponding to the four defect categories. Each method generates 1,000 images for each defect category and combines them as a whole training set with 4,000 images. Finally, for each method, we train a ResNet-34 \cite{resnet} on its own training set and evaluate on the shared test set. We repeat these experiments three times with different partitions of base sets and test sets to avoid bias.

The accuracies on the test set are shown in \cref{tab:class}. In this classification task simulating real-world defect inspection in industries, \name{} achieves the highest scores on all the partitions, with generally 10\% improvement than the runner-up. It means that our method serves as the most informative data augmentation technique for the downstream task.

%%%%%%%%%%%%%%%%%%%%
% CONCLUSION
%%%%%%%%%%%%%%%%%%%%

\section{Conclusion}
\label{sec:conclusion}
In this work, we propose the first few-shot defect image generation method \name{} which is capable of generating diverse defect images with high quality based on just a handful of defect samples. \name{} features its defect-aware residual blocks, which learn to produce reasonable defect masks and accordingly manipulate the object features. The highlight advantage is that it eases the transfer process from defect-free data to defect ones by delimiting the manipulation within the defect regions to focus solely on generating defects. Experiments on MVTec AD have proved the strong generation abilities of our method, as well as its benefits for downstream defect inspection tasks. We will discuss possible future works in \emph{supplementary material}.

\section*{Acknowledgements}
The work was supported by the National Science Foundation of China (62076162), and the Shanghai Municipal Science and Technology Major/Key Project, China (2021SHZDZX0102, 20511100300).

\bibliography{egbib}

\end{document}

% \maketitle
\begin{center}
    \textbf{\Large{Supplementary Material}}
\end{center}

As the supplementary material for \emph{Few-Shot Defect Image Generation via Defect-Aware Feature Manipulation}, we will list implementation details in \cref{sec:implementation}, show additional experimental results in \cref{sec:experiment}, and describe potential future works based on \name{} in \cref{sec:limit}. 

\section{Implementation Detail}
\label{sec:implementation}

\subsection{Network Structure}

We generally follow the original implementation of StyleGAN2 \cite{stylegan2} with ADA \cite{stylegan2-ada} provided by NVIDIA. For the details of the generator backbone and the discriminator, please refer to their official GitHub repository\footnote{\url{https://github.com/NVlabs/stylegan2-ada-pytorch}}.

We modify the aforementioned codes to implement our
\name{}, mainly including the defect mapping network, the defect-aware residual blocks and the defect matching discriminator.

\paragraph{Defect Mapping Network}

The defect mapping network has exactly the same structure with the mapping network of the backbone. It consists of two fully-connected layers, whose outputs have 512 channels.

\paragraph{Defect-aware Residual Block}

The defect-aware residual blocks share the same network structures (two convolutional layers) with the synthesis block at the same resolution. Besides, the first defect-aware residual block at resolution 64 is accompanied with a \emph{ToMask} module similar to \emph{ToRGB} of the backbone, with one output channel (mask) instead of three (RGB image).

\paragraph{Defect Matching Discriminator}

The defect matching discriminator generally has the identical network structure with the StyleGAN2 discriminator, which has discriminator blocks from resolution 256 to 8, and finally a discriminator epilogue at resolution 4. Yet, we modify the number of input channels of the first discriminator block from three to four because the defect matching discriminator takes the concatenated image and mask as its input. Also, reducing the number of channels of each convolutional layer to 1/4 eases training processes on limited data for most cases. 

\subsection{Hyperparameter Choice}

The choices of the new hyperparameters imported by additional modules of \name{} are specified in the main paper. For the hyperparameters used by StyleGAN2, we choose the default values. All the random codes $\bm{z}$ and modulation weights $\bm{w}$ have 512 dimensions, and the batch size is set to 32. For the optimization, we use Adam optimizer with learning rate $0.0025$, betas $(0, 0.99)$ and epsilon $10^{-8}$. The random seeds of all experiments are set to 0 by default.

\subsection{Training Protocol}

Following the implementation of StyleGAN2, we compose the codes of \name{} using PyTorch and are available at \url{https://github.com/Ldhlwh/DFMGAN}.

For the defect image generation experiments, we distribute the training processes on two GPUs and train for 400 kimgs. For the defect classification experiments, we train a ResNet-34 \cite{resnet} for each setting with learning rate $10^{-5}$, batch size 64, 50 epochs on one GPU.
\section{Additional Experiment}
\label{sec:experiment}

In \cref{sec:ablation}, we review several salient choices \wrt the architecture of \name{} by conducting ablation study on the object category of hazelnut. In \cref{sec:few-shot}, we challenge our model with even fewer defect samples (5-shot or 1-shot) provided for training in order to validate its strong generation ability. In \cref{sec:sig}, we do significance tests to check whether the improvements brought by \name{} are statistically significant.

To prove that our \name{} is generally applicable to few-shot defect image generation for various object/texture, and their defect categories, we conduct additional experiments on all the other categories of MVTec AD \cite{mvtecad} besides the object category \emph{hazelnut} in the main paper. These experiments share the same settings with those on \emph{hazelnut}, including the metrics, the baselines, and the two parts of experiments (\ie, defect image generation in \cref{sec:exp-dig} and defect classification in \cref{sec:exp-dc}).

\subsection{Ablation Study}
\label{sec:ablation}

\begin{figure*}[t]
  \centering
  \includegraphics[width=1.0\textwidth]{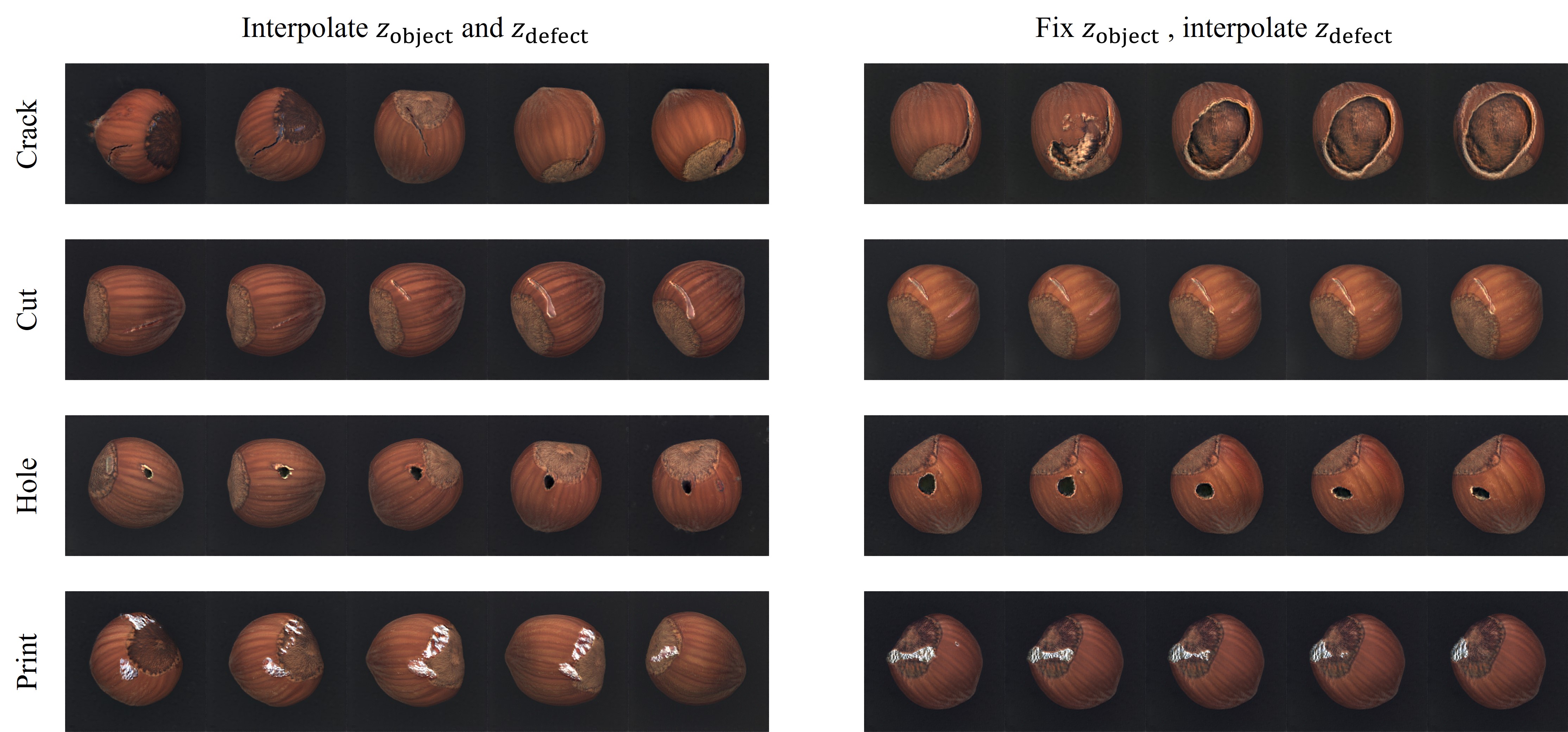}
  \caption{Examples of the generated defect images with interpolated random code(s).}
  \label{fig:itp}
\end{figure*}

\begin{table}[t]
  \centering
  \begin{tabular}{lrr}
    \hline
    Method/Variant & KID$\downarrow$ & LPIPS$\uparrow$ \\
    \hline
    ResBlock32 & 28.49 & 0.2126 \\
    ResBlock128 & 25.90 & 0.2214 \\
    ReplaceFeat & 21.44 & 0.2293 \\
    UnifiedDis & 69.44 & 0.2302 \\
    MSWI & 23.87 & 0.2225 \\
    NoMS & 21.58 & 0.2124 \\
    \hline
    \textbf{\name{}} & \textbf{20.78} & \textbf{0.2391} \\
    \hline
  \end{tabular}
  \caption{The results of the ablation study evaluating several variants of \name{}.}
  \label{tab:ablation}
\end{table}

We compare our model on the typical defect category \emph{hole} with the following variants: (1) \emph{ResBlock32}: the defect-aware residual blocks start from resolution 32 instead of 64; (2) \emph{ResBlock128}: the residual blocks start from resolution 128; (3) \emph{ReplaceFeat}: the defect features are used to replace the object features in the defect regions instead of summing up as residual features; (4) \emph{UnifiedDis}: train one unified discriminator instead of two, where the image features and the mask features are fused at middle level with the image branch inheriting the parameters of the discriminator $D$ trained in the first stage; (5) \emph{MSWI}: minimize $\Delta \bm{w} / \Delta \mathrm{image}$ instead of $\Delta \bm{w} / \Delta\mathrm{mask}$ for mode seeking loss; (6) \emph{NoMS}: remove the mode seeking loss. As shown in \cref{tab:ablation}, our \name{} has the best performance on both metrics. Also, except for UnifiedDis which remarkably changes the structure of our model, the other slightly modified variants all achieve relatively good performances, manifesting the robustness of \name{}.

Our \name{} takes two random codes $\zobject$ and $\zdefect$ as its input. To show that these codes respectively provide correct control to the generated defect images, we present interpolation examples in \cref{fig:itp}, where we either linearly interpolate both codes for full variations, or interpolate $\zdefect$ only for generating defects on the same object determined by the fixed $\zobject$. As expected, the object appearances are exclusively decided by $\zobject$, while the defects are mutually determined by $\zobject$ and $\zdefect$ with continuous variations. Also, since defects can continuously change with $\zdefect$ on the same object, it further improves the diversity of the generated defect images.

\subsection{Few-shot Generation}
\label{sec:few-shot}

\begin{table}[t]
  \centering
    \begin{tabular}{lrr|rr}
    \hline
    & \multicolumn{2}{c|}{5-shot} & \multicolumn{2}{c}{1-shot}  \\
    %\hline
    Method & KID$\downarrow$ & LPIPS$\uparrow$ & KID$\downarrow$ & LPIPS$\uparrow$  \\
    \hline
    Finetune & \textbf{16.26} & 0.1027 & 79.56 & 0.1097  \\
    DiffAug & 24.05 & 0.0335 & 61.23 & 0.1082  \\
    CDC & 57.06 & 0.1696 & 81.14 & 0.1861  \\
    \hline
    Crop\&Paste & - & 0.2242 & - & 0.2164 \\
    \hline
    SDGAN & 430.14 & 0.1917 & 466.84 & 0.1975 \\
    Defect-GAN & 88.90 & 0.2123 & 181.06 & 0.2075 \\
    \textbf{\name{}} & 21.88 & \textbf{0.2620} & \textbf{36.41} & \textbf{0.2554} \\
    \hline
  \end{tabular}
  \caption{The results of 5-shot and 1-shot defect image generation experiments, where we report KID$\times 10^3$@5k and clustered LPIPS@1k for each setting. We keep using the full \emph{hole} category (18 images) for a better estimation when calculating KIDs.}
  \label{tab:few-shot}
\end{table}

\begin{figure*}[t]
  \centering
  \includegraphics[width=1.0\textwidth]{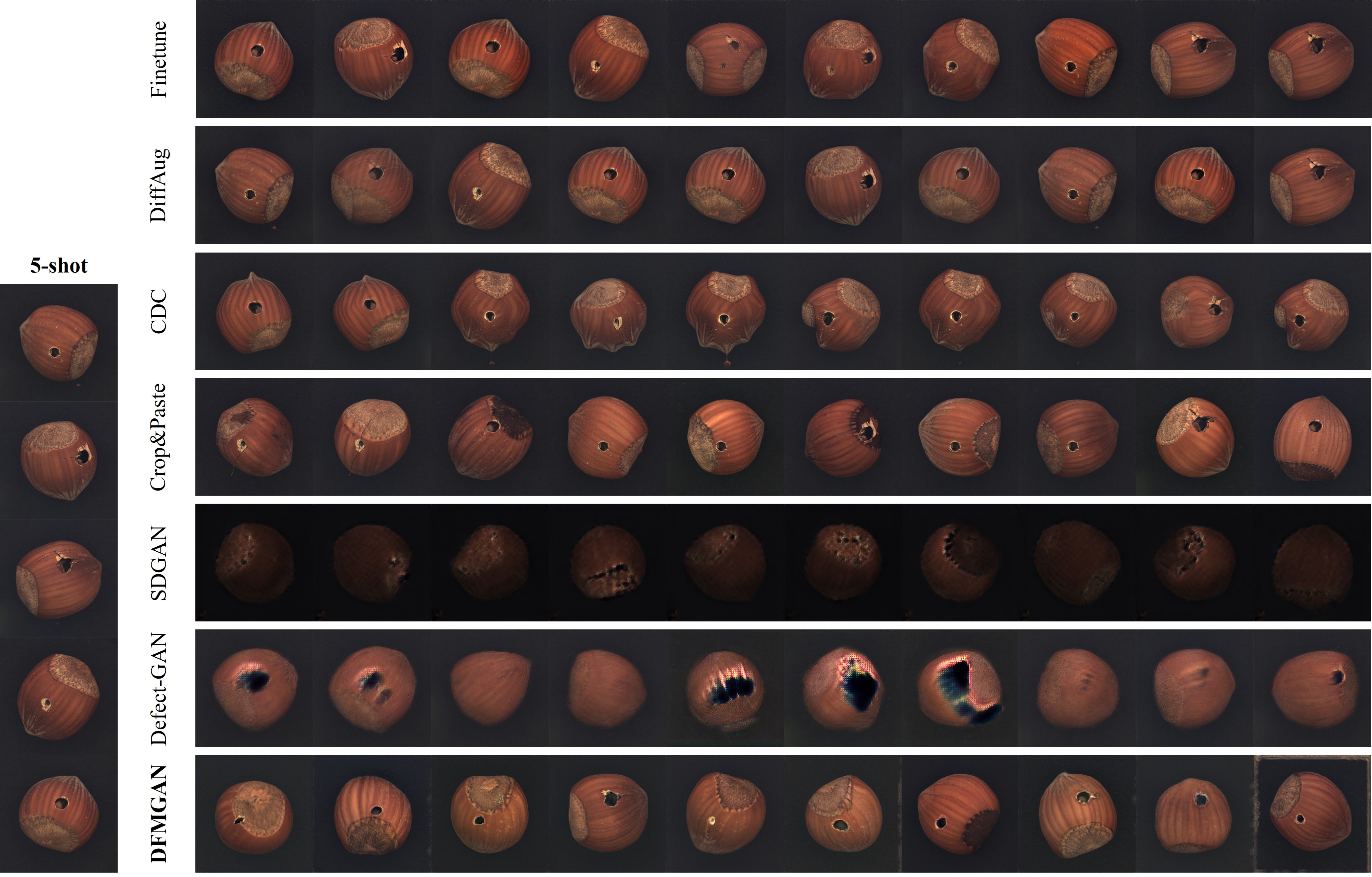}
  \caption{The dataset used for 5-shot defect image generation and examples of the generated defect images.}
  \label{fig:5-shot}
\end{figure*}
\begin{figure*}[t]
  \centering
  \includegraphics[width=1.0\textwidth]{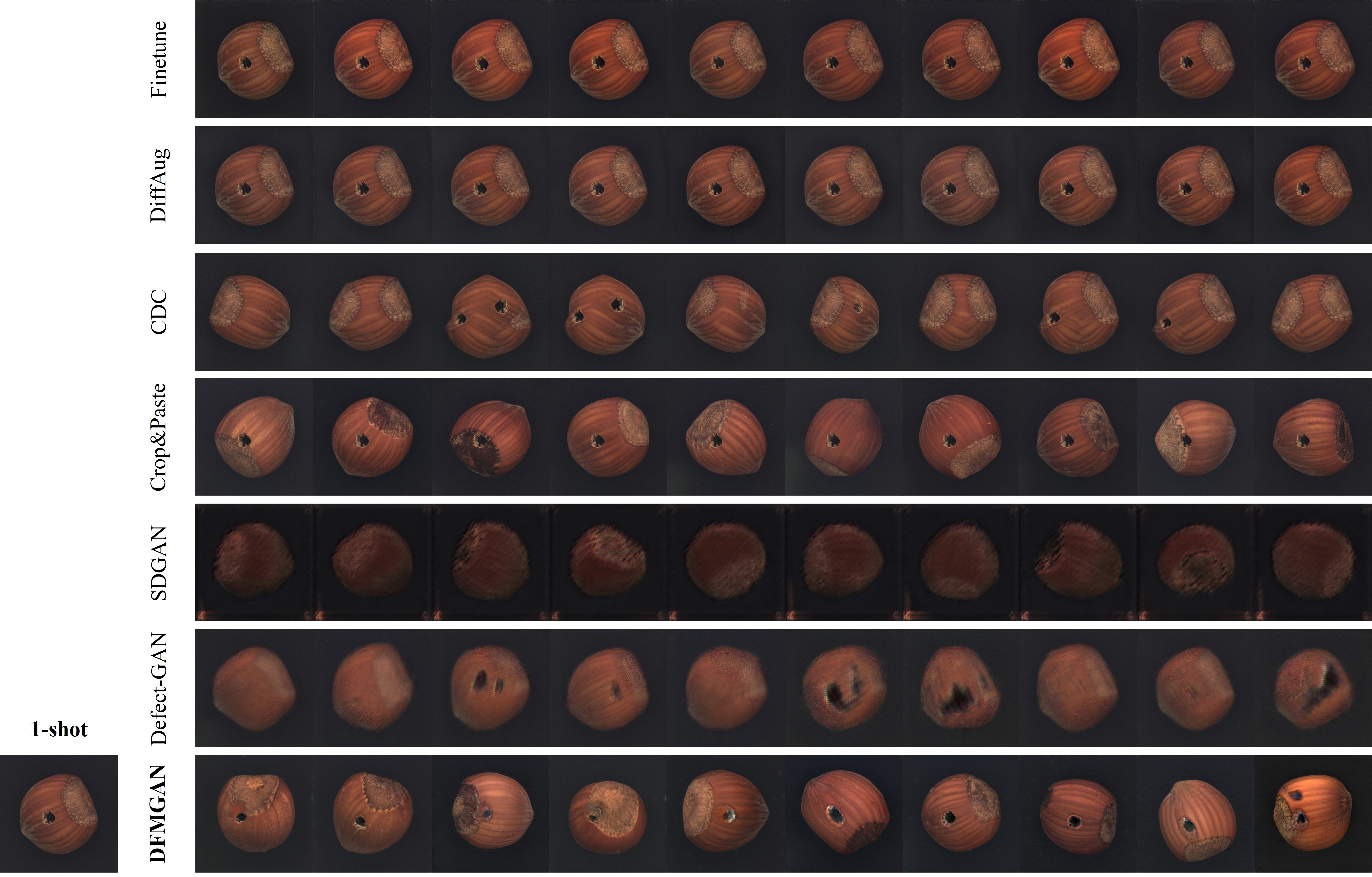}
  \caption{The dataset used for 1-shot defect image generation and examples of the generated defect images.}
  \label{fig:1-shot}
\end{figure*}

We redo the defect image generation experiments on the defect category \emph{hole} of object category \emph{hazelnut}, yet the training processes only utilize five or one randomly chosen defect images from the full dataset of 18 images. The quantitative and the qualitative results are shown in \cref{tab:few-shot} and \cref{fig:5-shot} (5-shot), \cref{fig:1-shot} (1-shot). Similar to the generation experiments using full dataset in the main paper, Finetune and DiffAug \cite{diffaug} keep repeating the dataset images, while CDC \cite{cross}, SDGAN \cite{sdgan} and Defect-GAN \cite{defectgan} cannot render realistic samples. The fact that Crop\&Paste \cite{nbr} does not generate new defect is particularly obvious in the 1-shot case (see \cref{fig:1-shot}), where all the generated samples exactly share the same (and the only) defect from the single dataset image. Our \name{} still achieves satisfying performance without much deterioration even in these extreme few-shot cases.

\subsection{Significance Test}
\label{sec:sig}

\begin{table}[t]
  \centering
  \begin{tabular}{lcc}
    \hline
    Method & KID$\downarrow$ & LPIPS$\uparrow$ \\
    \hline
    Defect-GAN & 37.54 $\pm$ 0.96 & 0.1898 $\pm$ 0.0012 \\
    \textbf{\name{}} & 21.08 $\pm$ 0.75 & 0.2396 $\pm$ 0.0056 \\
    \hline
  \end{tabular}
  \caption{The results of significance tests between \name{} and Defect-GAN, where we report Mean $\pm$ Std of ten times.}
  \label{tab:sig}
\end{table}

We perform significance tests between our \name{} and the strongest baseline Defect-GAN \cite{defectgan} on the defect category \emph{hole} of object category \emph{hazelnut}, where we repeat the experiments ten times using different random seeds. According to the quantitative results in \cref{tab:sig}, we conduct Welch's $t$-test on both metrics. The $p$ values on KID and clustered LPIPS are respectively $2.53 \times 10^{-18}$ and $2.23 \times 10^{-10}$, which demonstrates the statistically significant superiority of our method.

\subsection{Defect Image Generation}
\label{sec:exp-dig}

The quantitative results of the categories \emph{bottle}, \emph{cable}, \emph{capsule}, \emph{carpet}, \emph{grid}, \emph{leather}, \emph{metalnut}, \emph{pill}, \emph{screw}, \emph{tile}, \emph{toothbrush}, \emph{wood} and \emph{zipper} are respectively listed in \cref{tab:bottle,tab:cable,tab:capsule,tab:carpet,tab:grid,tab:leather,tab:metalnut,tab:pill,tab:screw,tab:tile,tab:toothbrush,tab:transistor,tab:wood,tab:zipper}, and their qualitative results are illustrated in \cref{fig:bottle,fig:cable,fig:capsule,fig:carpet,fig:grid,fig:leather,fig:metalnut,fig:pill,fig:screw,fig:tile,fig:toothbrush,fig:transistor,fig:wood,fig:zipper}. Samples generated by Finetune and DiffAug are accompanied with their closest dataset images as in our main paper.

\paragraph{Reality} Due to the following reasons, our proposed \name{} does not always achieve the best KID scores among the baselines. (1) Generic few-shot image generation methods including Finetune and DiffAug can generate images with nearly identical distribution with the dataset by overfit to the few given defect images. Hence these methods can reach very little KID scores. (2) In some cases where the defects are subtle (\ie the defect images highly resemble the defect-free ones), previous defect image generation methods SDGAN and Defect-GAN fail to render defects. They keep generating defect-free samples while still have good KID scores (\eg the category \emph{capsule} in \cref{tab:capsule,fig:capsule}, and the category \emph{pill} in \cref{tab:pill,fig:pill}). On the contrary, \name{} always try to generate defects despite of possible KID increment. Later the results of the defect classification experiments will prove that \name{} actually provides much more informative data augmentation than the baselines achieving better KID scores.

\paragraph{Diversity} For most of the cases (56 out of 73 including the four defect categories of \emph{hazelnut}), our model gets the highest clustered LPIPS scores, and in the other cases \name{} is generally either just marginally outperformed, or the non-generative method Crop\&Paste gains diversity by pasting defects off the objects (\eg the category \emph{screw} in \cref{tab:screw,fig:screw}). These quantitative results indicate the uniformly satisfying diversity of our proposed \name{}.

\subsection{Data Augmentation for Defect Classification}
\label{sec:exp-dc}

Following the data augmentation procedures for the defect classification tasks on \emph{hazelnut}, we conduct the same experiments on the other categories as well. The results are shown in \cref{tab:bottle-class,tab:cable-class,tab:capsule-class,tab:carpet-class,tab:grid-class,tab:leather-class,tab:metalnut-class,tab:pill-class,tab:screw-class,tab:tile-class,tab:transistor-class,tab:wood-class,tab:zipper-class}. Note that we omit the object category \emph{toothbrush} because it only has one defect category.

For 36 out of 42 cases (including the three partitions of \emph{hazelnut}), the datasets augmented by \name{} achieve the best accuracies, and in the other six cases our model ranks the second place. Although our model may not render dataset-like defect images as the overfitted Finetune or DiffAug do, the fairly realistic and highly diverse generated images still lead to the most informative data augmentation to enhance the downstream tasks. The results again prove that when designing data augmentation methods to enhance the defect inspection systems, it is desirable to generate slightly less realistic yet much more diverse defect images rather than excessively real images resembling the few training samples.

\section{Future Work}
\label{sec:limit}

Objectively speaking, despite the satisfying performances of our model, a few limitations still imply potential future researches. 

First, \name{} may not perform well when the defects induce significant changes of the object contours, including (1) destructive defects removing part of the objects (\eg, some severe cracks of hazelnuts) since our model is not well trained on filling the missing parts with reasonable background, and (2) additive defects extending the object areas (\eg some misplaced transistors) since our model needs to generate defects on the original background with non-object feature. 

Second, up to now we present our \name{} in an unconditional version, which learns to generate various types of defects one at a time. During the experiment period of this work, we actually tried a conditional version of \name{} where our model can be controlled to generate all types of defects given extra category codes. However, the conditional \name{} prone to yield sub-optimal performance on some of the defect categories while has good performances on the others. Besides, some baselines of this work (CDC, SDGAN) are unconditional. Due to the reasons above, we finally decided to make our \name{} unconditional in this work. Nevertheless, if these issues are settled, we suppose a conditional model sharing part of the information across defect categories might be desirable. We hope that our research, aiming at being a pivotal work, may give rise to future research effort in the area of defect image generation.
\bibliography{egbib}
\newpage
\newcommand{\threewidth}{0.8}
\newcommand{\singlefourwidth}{1.0}
\newcommand{\doublefourwidth}{0.9}
\newcommand{\fivewidth}{0.67}

%%%%% bottle
\begin{table*}[t]
  \centering
  \begin{tabular}{lrr|rr|rr}
    \hline
    Bottle & \multicolumn{2}{c|}{Broken Large} & \multicolumn{2}{c|}{Broken Small} & \multicolumn{2}{c}{Contamination} \\
    Method & KID$\downarrow$ & LPIPS$\uparrow$ & KID$\downarrow$ & LPIPS$\uparrow$ & KID$\downarrow$ & LPIPS$\uparrow$ \\
    \hline
    Finetune & 117.77 & 0.0565 & 138.50 & 0.0387 & 131.67 & 0.0548 \\
    DiffAug & \textbf{16.14} & 0.0327 & \textbf{23.61} & 0.0270 & \textbf{6.28} & 0.0294 \\
    CDC & 65.35 & 0.0429 & 114.47 & 0.0344 & 155.73 & 0.0444 \\
    \hline
    Crop\&Paste & - & 0.0401 & - & 0.0448 & - & 0.0479\\
    \hline
    SDGAN & 184.24 & 0.0688 & 201.35 & 0.0498 & 192.11 & 0.0483 \\
    Defect-GAN & 77.09 & 0.0593 & 59.18 & 0.0797 & 126.45 & 0.0693 \\
    \textbf{\name{}} & 59.74 & \textbf{0.1162} & 76.38 & \textbf{0.0854} & 76.59 & \textbf{0.1661} \\
    \hline
  \end{tabular}
  \caption{The results of the few-shot defect image generation experiments on object category \emph{bottle} with three defect categories \emph{broken large}, \emph{broken small} and \emph{contamination}.}
  \label{tab:bottle}
\end{table*}

\begin{table*}[t]
  \centering
  \begin{tabular}{lrrr}
    \hline
    Bottle & P1 Acc$\uparrow$ & P2 Acc$\uparrow$ & P3 Acc$\uparrow$ \\
    \hline
    Finetune & 37.21 & 41.86 & 41.86 \\
    DiffAug   & 48.84 & 44.18 & 53.49 \\
    CDC   & 44.19 & 37.21 & 34.88 \\
    \hline
    Crop\&Paste   & 53.49 & 51.16 & 53.49 \\
    \hline
    SDGAN   & 48.84 & 46.51 & 51.16 \\
    Defect-GAN   & 53.49 & 53.49 & 53.49 \\
    \textbf{\name{}} & \textbf{55.81} & \textbf{55.81} & \textbf{58.14} \\
    \hline
  \end{tabular}
  \caption{The results of the defect classification experiments on object category \emph{bottle}.}
  \label{tab:bottle-class}
\end{table*}

\begin{figure*}[t]
  \centering
  \includegraphics[width=\threewidth\linewidth]{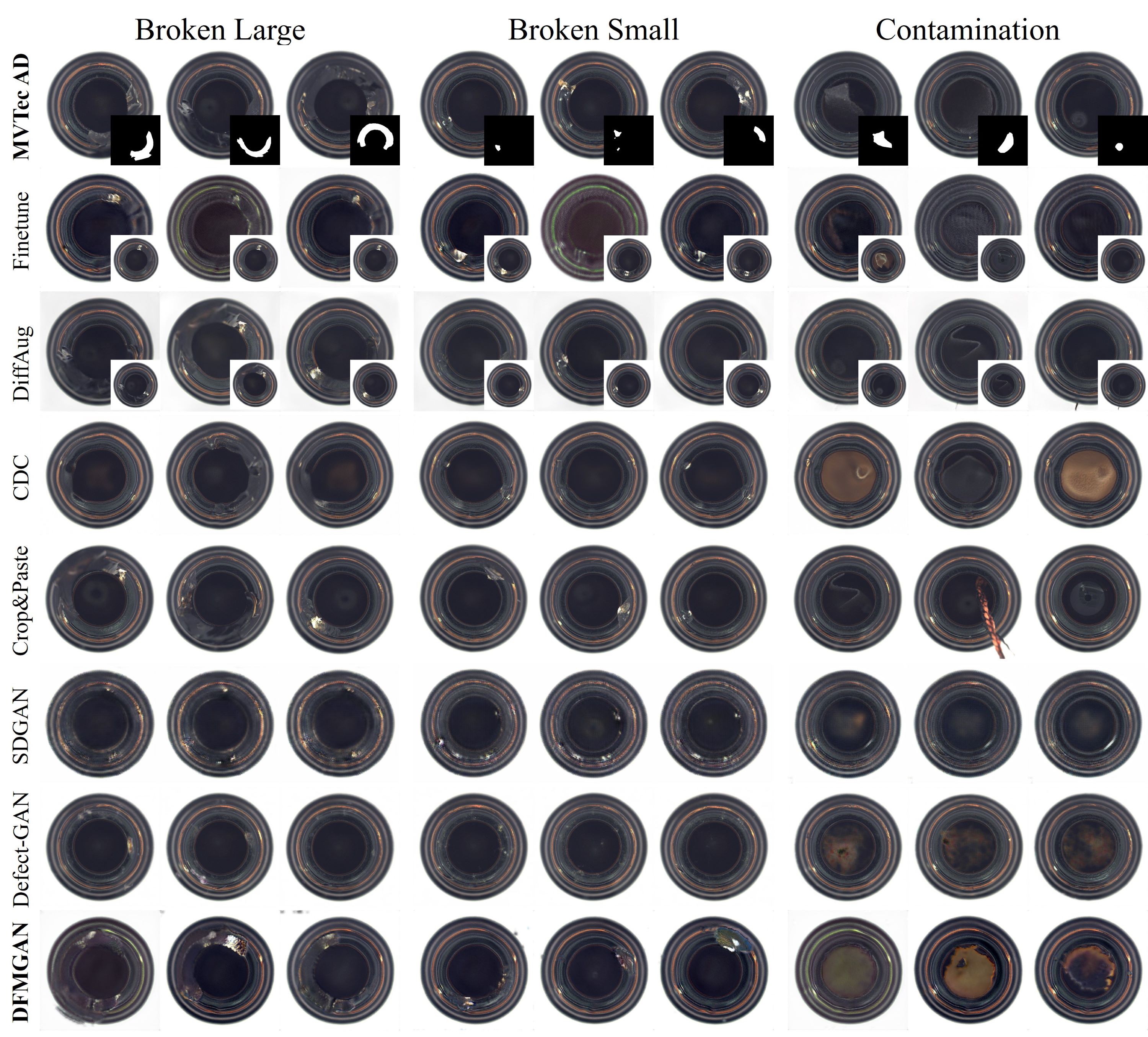}
   \caption{Examples of datasets (with masks) and generated defect images by different methods on object category \emph{bottle}. }
   \label{fig:bottle}
\end{figure*}

\clearpage

%%%%% cable
\begin{table*}
  \centering
  \begin{tabular}{lrr|rr|rr|rr}
    \hline
    Cable & \multicolumn{2}{c|}{Bent Wire} & \multicolumn{2}{c|}{Cable Swap} & \multicolumn{2}{c|}{Combined} & \multicolumn{2}{c}{Cut In. Insu.} \\
    Method & KID$\downarrow$ & LPIPS$\uparrow$ & KID$\downarrow$ & LPIPS$\uparrow$ & KID$\downarrow$ & LPIPS$\uparrow$ & KID$\downarrow$ & LPIPS$\uparrow$ \\
    \hline
    Finetune & \textbf{9.80} & 0.1155 & 5.83 & 0.1374 & \textbf{7.65} & 0.1364 & \textbf{17.63} & 0.1222 \\
    DiffAug   & 284.15 & 0.0449 & \textbf{5.58} & 0.0632 & 21.93 & 0.0531 & 79.18 & 0.1001 \\
    CDC   & 61.05 & 0.1944 & 54.34 & 0.1714 & 64.85 & 0.1910 & 53.94 & 0.1897 \\
    \hline
    Crop\&Paste   & - & 0.2693 & - & \textbf{0.2502} & - & 0.2726 & - & 0.2354 \\
    \hline
    SDGAN   & 456.99 & 0.1407 & 518.76 & 0.2034 & 445.38 & 0.1931 & 515.56 & 0.1966 \\
    Defect-GAN   & 139.87 & 0.2181 & 99.03 & 0.2064 & 68.18 & 0.2126 & 259.76 & 0.2236 \\
    \textbf{\name{}} & 48.95 & \textbf{0.2782} & 33.60 & 0.2120 & 50.47 & \textbf{0.2745} & 51.92 & \textbf{0.2487} \\
    \hline
  \end{tabular}
  \begin{tabular}{lrr|rr|rr|rr}
    \hline
    Cable & \multicolumn{2}{c|}{Cut Out. Insu.} & \multicolumn{2}{c|}{Missing Cable} & \multicolumn{2}{c|}{Missing Wire} & \multicolumn{2}{c}{Poke Insu.}\\
    Method & KID$\downarrow$ & LPIPS$\uparrow$ & KID$\downarrow$ & LPIPS$\uparrow$ & KID$\downarrow$ & LPIPS$\uparrow$ & KID$\downarrow$ & LPIPS$\uparrow$ \\
    \hline
    Finetune & \textbf{14.58} & 0.1367 & \textbf{1.52} & 0.1120 & \textbf{3.05} & 0.1258 & \textbf{15.09} & 0.1252 \\
    DiffAug   & 117.19 & 0.0894 & 2.67 & 0.0543 & 7.45 & 0.0495 & 46.78 & 0.0633 \\
    CDC   & 93.05 & 0.2061 & 59.72 & 0.1718 & 46.21 & 0.1723 & 52.23 & 0.1985 \\
    \hline
    Crop\&Paste   & - & \textbf{0.2604} & - & \textbf{0.2519} & - & \textbf{0.2355} & - & \textbf{0.2491} \\
    \hline
    SDGAN   & 482.60 & 0.2194 & 423.80 & 0.1707 & 578.84 & 0.2086 & 634.73 & 0.1936 \\
    Defect-GAN   & 108.44 & 0.2164 & 174.06 & 0.2299 & 37.91 & 0.2167 & 200.85 & 0.2199 \\
    \textbf{\name{}} & 90.97 & 0.2599 & 60.09 & 0.2401 & 44.21 & \textbf{0.2355} & 51.43 & 0.2416 \\
    \hline
  \end{tabular}
  \caption{The results of the few-shot defect image generation experiments on object category \emph{cable} with eight defect categories \emph{bent wire}, \emph{cable swap}, \emph{combined}, \emph{cut inner insulation}, \emph{cut outer insulation}, \emph{missing cable}, \emph{missing wire} and \emph{poke insulation}.}
    \label{tab:cable}
\end{table*}

\begin{table*}[t]
  \centering
  \begin{tabular}{lrrr}
    \hline
    Cable & P1 Acc$\uparrow$ & P2 Acc$\uparrow$ & P3 Acc$\uparrow$ \\
    \hline
    Finetune & 39.06 & 39.06 & 40.63 \\
    DiffAug   & 25.00 & 23.44 & 15.63 \\
    CDC   & 40.63 & 43.75 & 32.81 \\
    \hline
    Crop\&Paste   & 39.06 & 28.13 & 31.25 \\
    \hline
    SDGAN   & 14.06 & 23.44 & 28.13 \\
    Defect-GAN   & 28.13 & 20.31 & 15.63 \\
    \textbf{\name{}} & \textbf{45.31} & \textbf{46.88} & \textbf{43.75} \\
    \hline
  \end{tabular}
    \caption{The results of the defect classification experiments on object category \emph{cable}.}
    \label{tab:cable-class}
\end{table*}

\clearpage

\begin{figure*}[t]
  \centering
  \includegraphics[width=\doublefourwidth\linewidth]{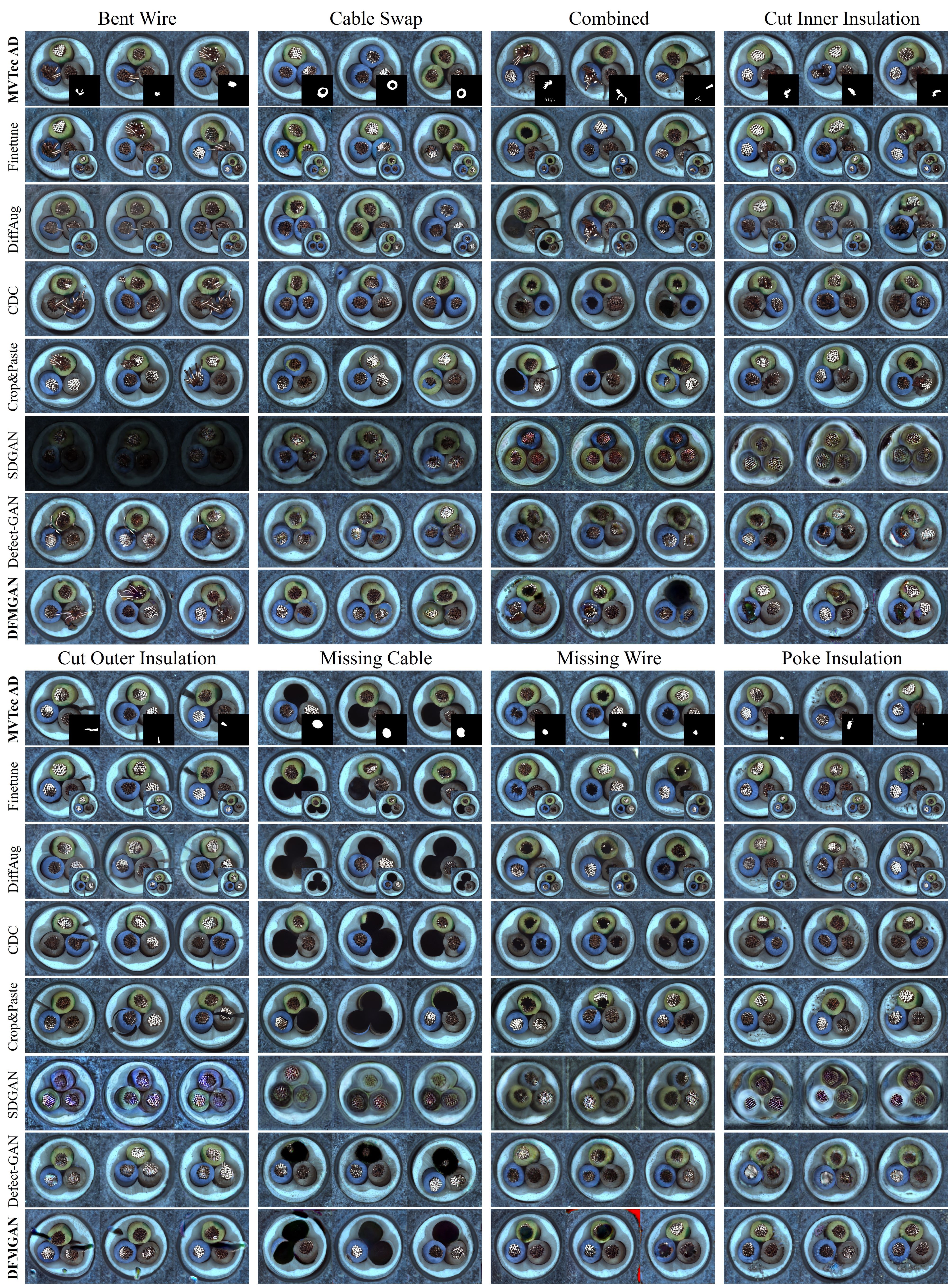}
   \caption{Examples of datasets (with masks) and generated defect images by different methods on object category \emph{cable}. }
   \label{fig:cable}
\end{figure*}

\clearpage

%%%%% capsule
\begin{table*}
  \centering
  \begin{tabular}{lrr|rr|rr|rr|rr}
    \hline
    Capsule & \multicolumn{2}{c|}{Crack} & \multicolumn{2}{c|}{Faulty Imprint} & \multicolumn{2}{c|}{Poke} & \multicolumn{2}{c|}{Scratch} & \multicolumn{2}{c}{Squeeze} \\
    Method & KID$\downarrow$ & LPIPS$\uparrow$ & KID$\downarrow$ & LPIPS$\uparrow$ & KID$\downarrow$ & LPIPS$\uparrow$ & KID$\downarrow$ & LPIPS$\uparrow$ & KID$\downarrow$ & LPIPS$\uparrow$ \\
    \hline
    Finetune & 19.52 & 0.0345 & 20.70 & 0.0357 & 18.71 & 0.0252 & 25.65 & 0.0263 & 31.29 & 0.0473 \\
    DiffAug   & 7.62 & 0.0301 & 16.07 & 0.0247 & 21.18 & 0.0252 & 17.77 & 0.0252 & \textbf{30.70} & 0.0385 \\
    CDC   & 48.35 & 0.0795 & 39.16 & 0.0604 & 36.33 & 0.0530 & 34.34 & 0.0509 & 131.80 & 0.0750 \\
    \hline
    Crop\&Paste   & - & 0.0484 & - & 0.0373 & - & 0.0483 & - & 0.0369 & - & 0.0604  \\
    \hline
    SDGAN   & 39.40 & 0.0265 & 37.80 & 0.0252 & 34.26 & 0.0289 & 39.54 & 0.0227 & 86.44 & 0.0465  \\
    Defect-GAN   & \textbf{3.27} & 0.0416 & \textbf{6.93} & 0.0371 & \textbf{7.13} & 0.0422 & \textbf{5.27} & 0.0307 & 92.63 & 0.0660 \\
    \textbf{\name{}} & 30.23 & \textbf{0.1183} & 49.65 & \textbf{0.1087} & 30.81 & \textbf{0.0614} & 28.69 & \textbf{0.1075} & 63.79 & \textbf{0.1286} \\
    \hline
  \end{tabular}
    \caption{The results of the few-shot defect image generation experiments on object category \emph{capsule} with five defect categories \emph{crack}, \emph{faulty imprint}, \emph{poke}, \emph{scratch} and \emph{squeeze}.}
    \label{tab:capsule}
\end{table*}

\begin{table*}[t]
  \centering
  \begin{tabular}{lrrr}
    \hline
    Capsule & P1 Acc$\uparrow$ & P2 Acc$\uparrow$ & P3 Acc$\uparrow$ \\
    \hline
    Finetune & 37.33 & 33.33 & 30.67 \\
    DiffAug   & 38.67 & \textbf{36.00} & 29.33 \\
    CDC   & 34.67 & 26.67 & 25.33 \\
    \hline
    Crop\&Paste   & 37.33 & 32.00 & 29.33 \\
    \hline
    SDGAN   & 34.67 & 29.33 & 26.67 \\
    Defect-GAN   & 30.67 & 33.33 & 32.00 \\
    \textbf{\name{}} & \textbf{41.33} & \textbf{36.00} & \textbf{34.37} \\
    \hline
  \end{tabular}
  \caption{The results of the defect classification experiments on object category \emph{capsule}.}
  \label{tab:capsule-class}
\end{table*}

\clearpage

% vertical * 5
\begin{figure*}[t]
  \centering
  \includegraphics[angle=-90, width=\fivewidth \linewidth]{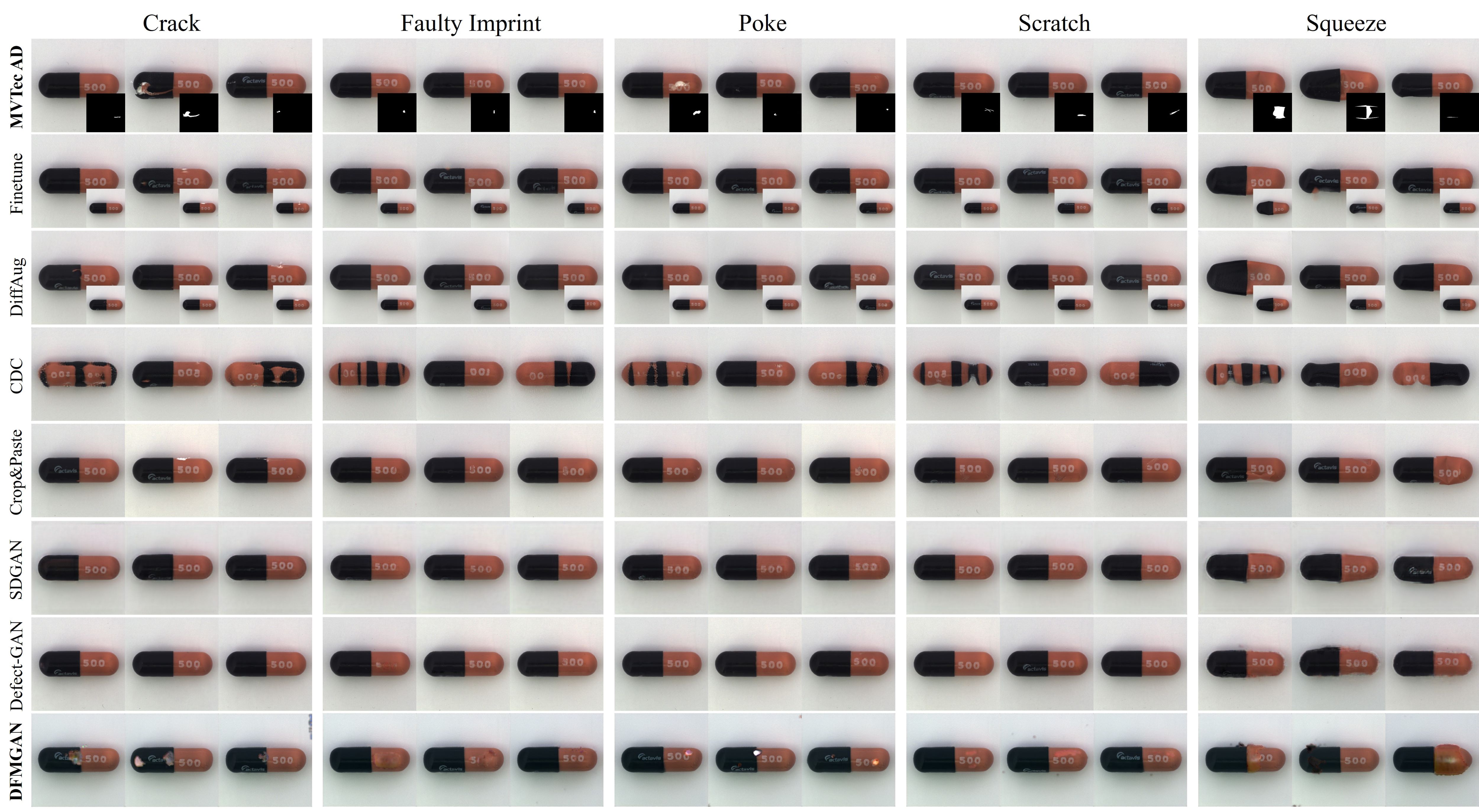}
   \caption{Examples of datasets (with masks) and generated defect images by different methods on object category \emph{capsule}. }
   \label{fig:capsule}
\end{figure*}

\clearpage

%%%%% carpet
\begin{table*}
  \centering
  \begin{tabular}{lrr|rr|rr|rr|rr}
    \hline
    Carpet & \multicolumn{2}{c|}{Color} & \multicolumn{2}{c|}{Cut} & \multicolumn{2}{c|}{Hole} & \multicolumn{2}{c|}{Metal Contam.} & \multicolumn{2}{c}{Thread} \\
    Method & KID$\downarrow$ & LPIPS$\uparrow$ & KID$\downarrow$ & LPIPS$\uparrow$ & KID$\downarrow$ & LPIPS$\uparrow$ & KID$\downarrow$ & LPIPS$\uparrow$ & KID$\downarrow$ & LPIPS$\uparrow$ \\
    \hline
    Finetune & \textbf{3.06} & 0.0806 & \textbf{4.46 }& 0.0902 & \textbf{2.94} & 0.0779 & 23.43 & 0.0776 & \textbf{16.65} & 0.0811 \\
    DiffAug   & 15.73 & 0.0562 & 31.46 & 0.0703 & 42.12 & 0.0586 & 19.91 & 0.0438 & 30.90 & 0.0503 \\
    CDC   & 22.02 & 0.0451 & 38.25 & 0.0408 & 21.15 & 0.0135 & 35.01 & 0.0099 & 47.90 & 0.0200 \\
    \hline
    Crop\&Paste   & - & 0.1100 & - & 0.1103 & - & 0.1084 & - & \textbf{0.1057} & - & 0.1161  \\
    \hline
    SDGAN   & 55.04 & 0.1123 & 55.64 & 0.1249 & 64.42 & 0.0873 & 49.68 & 0.0844 & 75.26 & 0.1206 \\
    Defect-GAN   & 26.00 & \textbf{0.1270} & 42.15 & 0.1254 & 71.30 & 0.1120 & 53.23 & 0.0975 & 44.19 & 0.1248 \\
    \textbf{\name{}} & 15.87 & 0.1236 & 29.87 & \textbf{0.1358} & 34.70 & \textbf{0.1157} & \textbf{19.89} & 0.1012 & 25.36 & \textbf{0.1648} \\
    \hline
  \end{tabular}
  \caption{The results of the few-shot defect image generation experiments on texture category \emph{carpet} with five defect categories \emph{color}, \emph{cut}, \emph{hole}, \emph{metal contamination} and \emph{thread}.}
  \label{tab:carpet}
\end{table*}

\begin{table*}[t]
  \centering
  \begin{tabular}{lrrr}
    \hline
    Carpet & P1 Acc$\uparrow$ & P2 Acc$\uparrow$ & P3 Acc$\uparrow$ \\
    \hline
    Finetune & 37.10 & 25.81 & 33.87 \\
    DiffAug   & 38.70 & 32.26 & 35.48 \\
    CDC   & 25.81 & 22.58 & 27.42 \\
    \hline
    Crop\&Paste   & 24.19 & 25.81 & 33.87 \\
    \hline
    SDGAN   & 20.97 & 24.19 & 19.35 \\
    Defect-GAN   & 35.48 & 22.58 & 29.03 \\
    \textbf{\name{}} & \textbf{46.77} & \textbf{46.77} & \textbf{48.39} \\
    \hline
  \end{tabular}
    \caption{The results of the defect classification experiments on texture category \emph{carpet}.}
      \label{tab:carpet-class}
\end{table*}

\clearpage

% vertical * 5
\begin{figure*}[t]
  \centering
  \includegraphics[angle=-90, width=\fivewidth \linewidth]{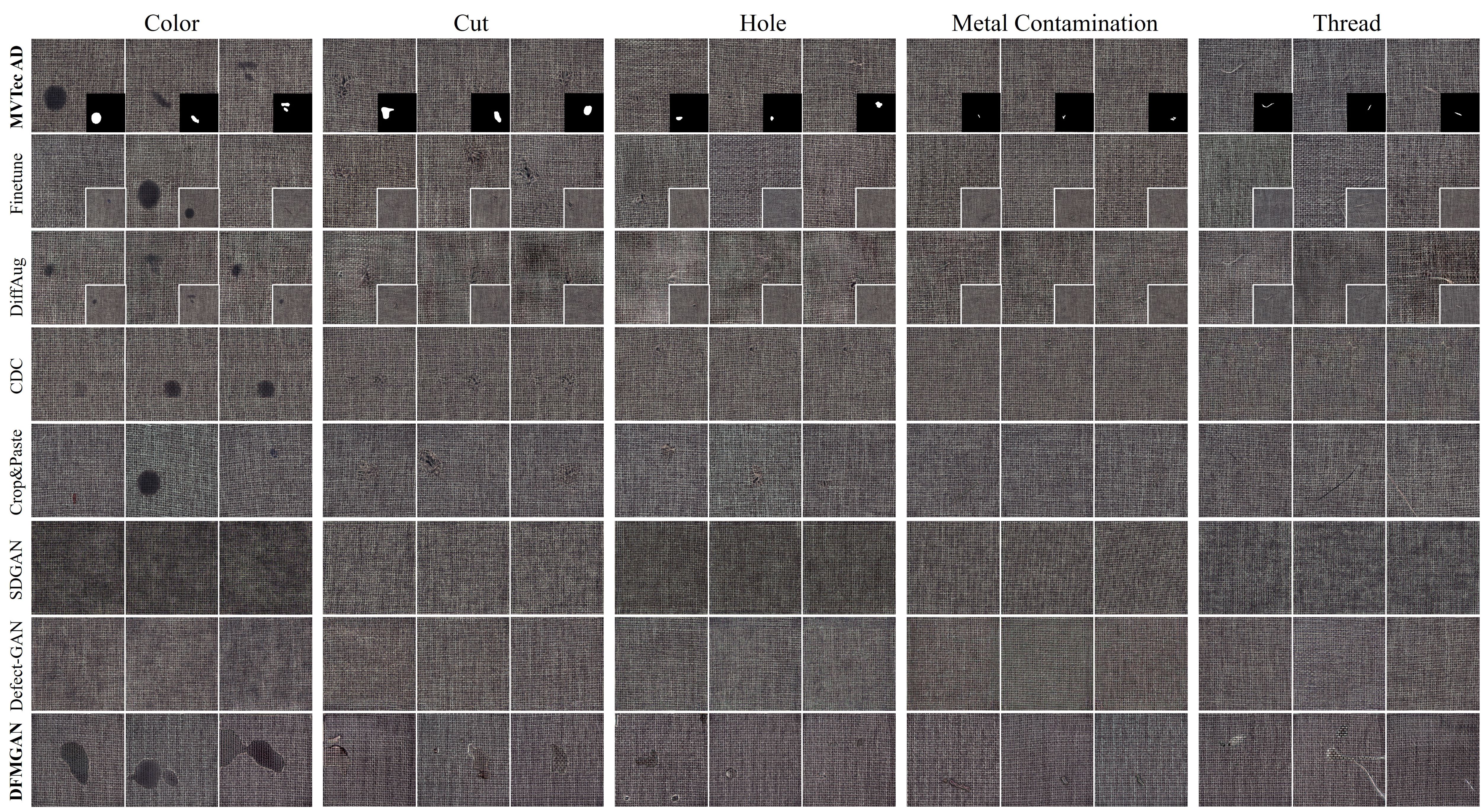}
   \caption{Examples of datasets (with masks) and generated defect images by different methods on texture category \emph{carpet}.   }
   \label{fig:carpet}
\end{figure*}

\clearpage

%%%%% grid
\begin{table*}
  \centering
  \begin{tabular}{lrr|rr|rr|rr|rr}
    \hline
    Grid & \multicolumn{2}{c|}{Bent} & \multicolumn{2}{c|}{Broken} & \multicolumn{2}{c|}{Glue} & \multicolumn{2}{c|}{Metal Contam.} & \multicolumn{2}{c}{Thread} \\
    Method & KID$\downarrow$ & LPIPS$\uparrow$ & KID$\downarrow$ & LPIPS$\uparrow$ & KID$\downarrow$ & LPIPS$\uparrow$ & KID$\downarrow$ & LPIPS$\uparrow$ & KID$\downarrow$ & LPIPS$\uparrow$ \\
    \hline
    Finetune & 19.90 & 0.0753 & \textbf{16.49} & 0.0824 & \textbf{16.69} & 0.0821 & 30.37 & 0.0844 & 7.65 & 0.0796 \\
    DiffAug   & 19.34 & 0.0583 & 99.01 & 0.0545 & 134.56 & 0.0575 & 186.43 & 0.0773 & \textbf{6.51} & 0.0627 \\
    CDC   & 21.40 & 0.0542 & 44.23 & 0.0617 & 23.14 & 0.0580 & \textbf{26.16} & 0.0637 & 104.44 & 0.0873 \\
    \hline
    Crop\&Paste   & - & \textbf{0.1074} & - & 0.1189 & - & 0.0994 & - & \textbf{0.1290} & - & 0.1217  \\
    \hline
    SDGAN   & 83.77 & 0.0937 & 57.25 & 0.1072 & 80.37 & 0.0955 & 83.65 & 0.0941 & 95.68 & 0.1235 \\
    Defect-GAN   & \textbf{8.36} & 0.1066 & 33.20 & 0.1376 & 134.56 & 0.0575 & 77.78 & 0.1228 & 84.27 & 0.1485 \\
    \textbf{\name{}} & 107.62 & 0.1069 & 117.12 & \textbf{0.1505} & 131.42 & \textbf{0.1259} & 60.67 & 0.0960 & 88.39 & \textbf{0.1624} \\
    \hline
  \end{tabular}
    \caption{The results of the few-shot defect image generation experiments on texture category \emph{grid} with five defect categories \emph{bent}, \emph{broken}, \emph{glue}, \emph{metal contamination} and \emph{thread}.}
    \label{tab:grid}
\end{table*}

\begin{table*}[t]
  \centering
  \begin{tabular}{lrrr}
    \hline
    Grid & P1 Acc$\uparrow$ & P2 Acc$\uparrow$ & P3 Acc$\uparrow$ \\
    \hline
    Finetune & 35.00 & 27.50 & 30.00 \\
    DiffAug   & 27.50 & 30.00 & 27.50 \\
    CDC   & 40.00 & 35.00 & 32.50 \\
    \hline
    Crop\&Paste   & 27.50 & 27.50 & 30.00 \\
    \hline
    SDGAN   & 30.00 & 27.50 & 35.00 \\
    Defect-GAN   & 25.00 & 22.50 & 35.00 \\
    \textbf{\name{}} & \textbf{47.50} & \textbf{37.50} & \textbf{37.50} \\
    \hline
  \end{tabular}
    \caption{The results of the defect classification experiments on texture category \emph{grid}.}
    \label{tab:grid-class}
\end{table*}

\clearpage

% vertical * 5
\begin{figure*}[t]
  \centering
  \includegraphics[angle=-90, width=\fivewidth \linewidth]{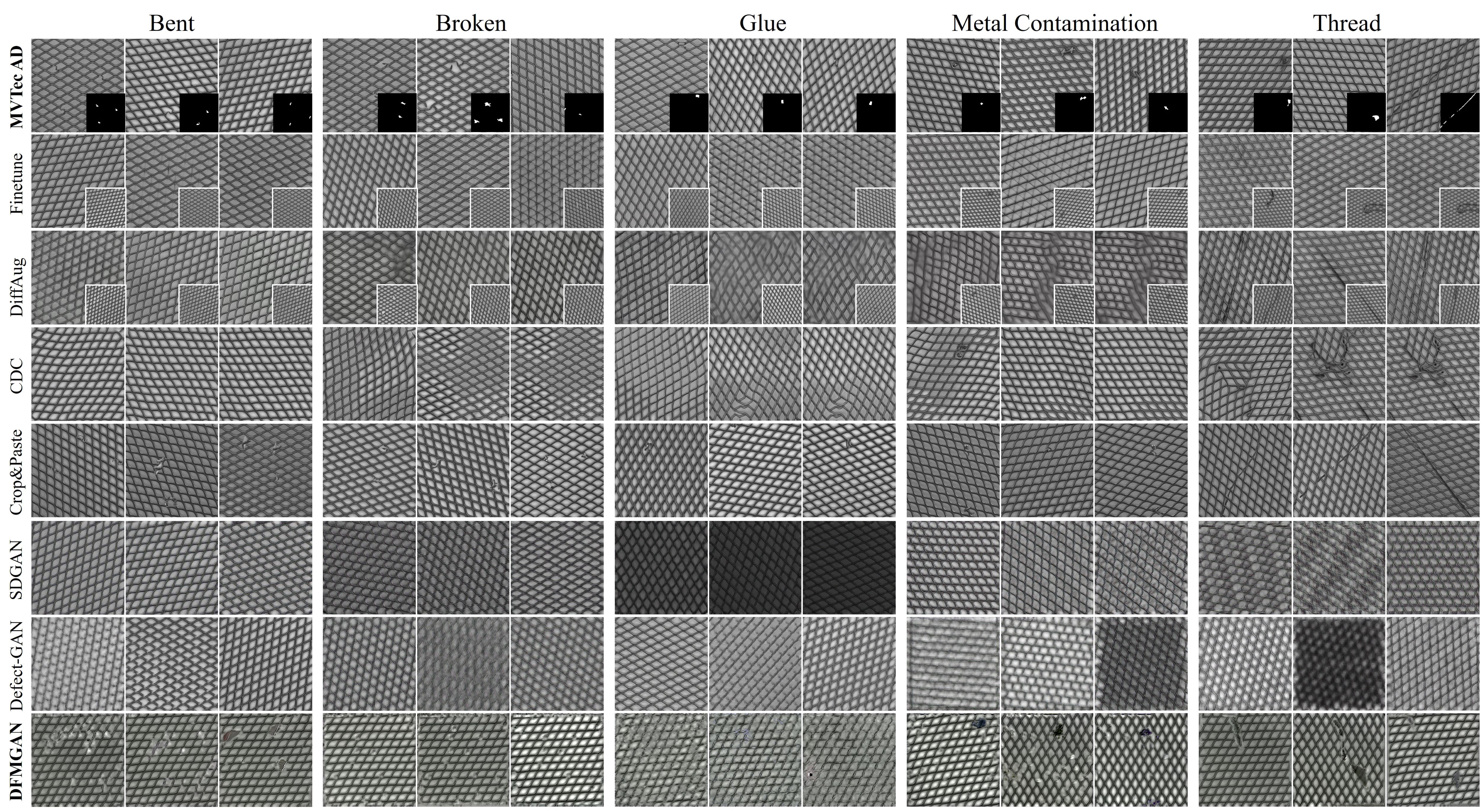}
   \caption{Examples of datasets (with masks) and generated defect images by different methods on texture category \emph{grid}.   }
   \label{fig:grid}
\end{figure*}

\clearpage

%%%%% leather
\begin{table*}
  \centering
  \begin{tabular}{lrr|rr|rr|rr|rr}
    \hline
    Leather & \multicolumn{2}{c|}{Color} & \multicolumn{2}{c|}{Cut} & \multicolumn{2}{c|}{Fold} & \multicolumn{2}{c|}{Glue} & \multicolumn{2}{c}{Poke} \\
    Method & KID$\downarrow$ & LPIPS$\uparrow$ & KID$\downarrow$ & LPIPS$\uparrow$ & KID$\downarrow$ & LPIPS$\uparrow$ & KID$\downarrow$ & LPIPS$\uparrow$ & KID$\downarrow$ & LPIPS$\uparrow$ \\
    \hline
    Finetune & 120.19 & 0.1055 & 49.59 & 0.1108 & 21.87 & 0.1013 & 45.22 & 0.0944 & 48.80 & 0.0968 \\
    DiffAug   & 113.81 & 0.0229 & \textbf{18.61} & 0.0683 & 37.03 & 0.0720 & \textbf{28.87} & 0.0648 & 42.26 & 0.0525 \\
    CDC   & \textbf{43.71} & 0.0524 & 25.45 & 0.0646 & \textbf{19.04} & 0.0750 & 41.83 & 0.0658 & \textbf{12.74} & 0.0773 \\
    \hline
    Crop\&Paste   & - & 0.1412 & - & \textbf{0.1366} & - & 0.1326 & - & 0.1345 & - & 0.1395 \\
    \hline
    SDGAN   & 548.64 & 0.1145 & 539.91 & 0.0941 & 450.63 & 0.1034 & 558.14 & 0.1324 & 536.24 & 0.1315 \\
    Defect-GAN   & 78.83 & 0.1368 & 120.56 & 0.1349 & 84.97 & 0.1375 & 182.03 & 0.1134 & 98.90 & 0.1499 \\
    \textbf{\name{}} & 59.17 & \textbf{0.1418} & 51.73 & 0.1283 & 141.34 & \textbf{0.2373} & 39.69 & \textbf{0.1520} & 87.32 & \textbf{0.1679} \\
    \hline
  \end{tabular}
    \caption{The results of the few-shot defect image generation experiments on texture category \emph{leather} with five defect categories \emph{color}, \emph{cut}, \emph{fold}, \emph{glue} and \emph{poke}.}
    \label{tab:leather}
\end{table*}

\begin{table*}[t]
  \centering
  \begin{tabular}{lrrr}
    \hline
    Leather & P1 Acc$\uparrow$ & P2 Acc$\uparrow$ & P3 Acc$\uparrow$ \\
    \hline
    Finetune & 46.03 & 49.21 & \textbf{47.02} \\
    DiffAug   & 41.27 & 39.68 & 41.27 \\
    CDC   & 39.68 & \textbf{58.73} & 31.74 \\
    \hline
    Crop\&Paste   & 31.74 & 36.51 & 34.92 \\
    \hline
    SDGAN   & 33.33 & 38.10 & 42.86 \\
    Defect-GAN   & 42.86 & 44.44 & 39.68 \\
    \textbf{\name{}} & \textbf{49.20} & 53.97 & 46.03 \\
    \hline
  \end{tabular}
    \caption{The results of the defect classification experiments on texture category \emph{leather}.}
    \label{tab:leather-class}
\end{table*}

\clearpage

% vertical * 5
\begin{figure*}[t]
  \centering
  \includegraphics[angle=-90, width=\fivewidth \linewidth]{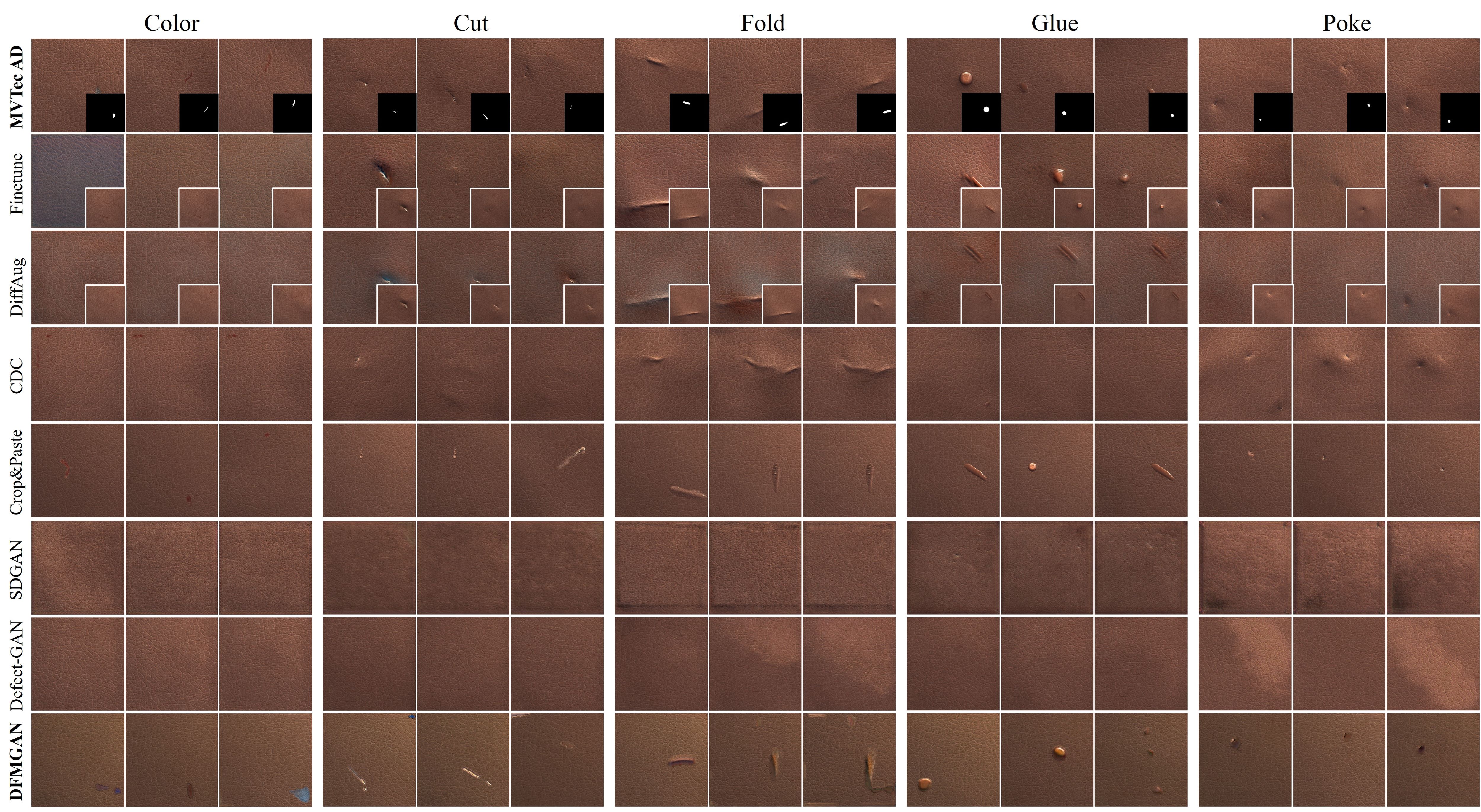}
   \caption{Examples of datasets (with masks) and generated defect images by different methods on texture category \emph{leather}.   }
   \label{fig:leather}
\end{figure*}

\clearpage

%%%%% metalnut
\begin{table*}[t]
  \centering
  \begin{tabular}{lrr|rr|rr|rr}
    \hline
    Metalnut & \multicolumn{2}{c|}{Bent} & \multicolumn{2}{c|}{Color} & \multicolumn{2}{c|}{Flip} & \multicolumn{2}{c}{Scratch} \\
    Method & KID$\downarrow$ & LPIPS$\uparrow$ & KID$\downarrow$ & LPIPS$\uparrow$ & KID$\downarrow$ & LPIPS$\uparrow$ & KID$\downarrow$ & LPIPS$\uparrow$ \\
    \hline
    Finetune & \textbf{13.35} & 0.2813 & \textbf{20.21} & 0.3153 & \textbf{16.25} & 0.2788 & \textbf{29.26} & 0.3142 \\
    DiffAug   & 21.65 & 0.2771 & 24.63 & 0.2939 & 24.70 & 0.2810 & 30.56 & 0.2943 \\
    CDC   & 175.10 & 0.0361 & 181.05 & 0.0468 & 167.88 & 0.0342 & 225.13 & 0.0412 \\
    \hline
    Crop\&Paste   & - & 0.1554 & - & 0.1529 & - & 0.1196 & - & 0.1805 \\
    \hline
    SDGAN   & 202.28 & 0.2892 & 248.49 & 0.2835 & 358.21 & 0.2736 & 160.27 & 0.2821 \\
    Defect-GAN   & 55.94 & 0.3058 & 44.83 & 0.3138 & 148.86 & 0.2836 & 56.29 & 0.3063 \\
    \textbf{\name{}} & 34.14 & \textbf{0.3153} & 35.72 & \textbf{0.3326} & 67.66 & \textbf{0.2919} & 38.65 & \textbf{0.3315} \\
    \hline
  \end{tabular}
    \caption{The results of the few-shot defect image generation experiments on object category \emph{metalnut} with four defect categories \emph{bent}, \emph{color}, \emph{flip} and \emph{scratch}.}
    \label{tab:metalnut}
\end{table*}

\begin{table*}[t]
  \centering
  \begin{tabular}{lrrr}
    \hline
    Metalnut & P1 Acc$\uparrow$ & P2 Acc$\uparrow$ & P3 Acc$\uparrow$ \\
    \hline
    Finetune & 60.94 & 59.37 & 60.94 \\
    DiffAug   & 56.25 & 57.81 & 62.50 \\
    CDC   & 40.63 & 48.44 & 56.25 \\
    \hline
    Crop\&Paste   & 57.81 & 59.37 & 62.50 \\
    \hline
    SDGAN   & 51.56 & 28.13 & 53.13 \\
    Defect-GAN   & 54.69 & 54.69 & 60.94 \\
    \textbf{\name{}} & \textbf{64.06} & \textbf{60.94} & \textbf{68.75} \\
    \hline
  \end{tabular}
    \caption{The results of the defect classification experiments on object category \emph{metalnut}.}
    \label{tab:metalnut-class}
\end{table*}

\begin{figure*}[t]
  \centering
  \includegraphics[width=\singlefourwidth\linewidth]{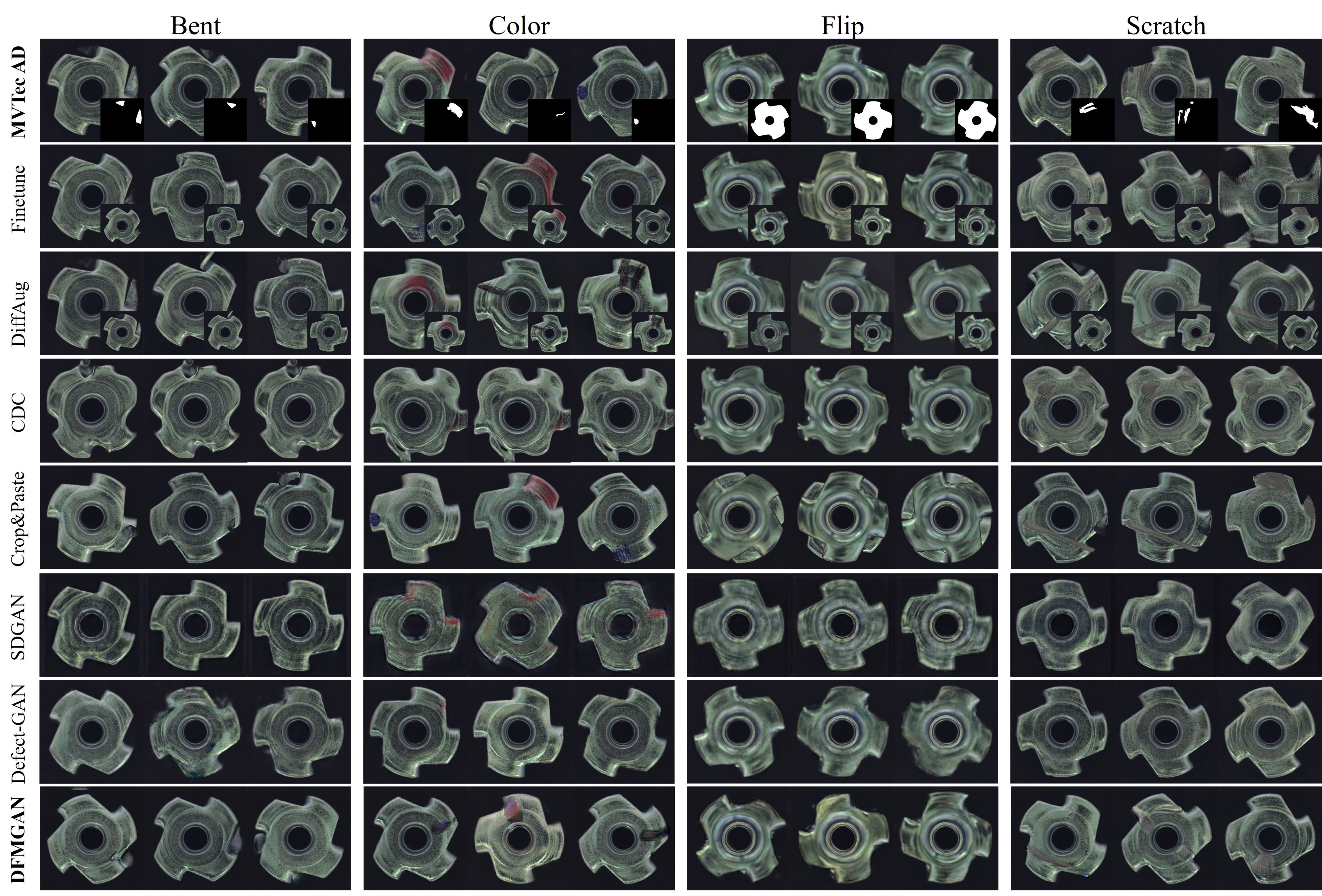}
   \caption{Examples of datasets (with masks) and generated defect images by different methods on object category \emph{metalnut}. }
   \label{fig:metalnut}
\end{figure*}

\clearpage

%%%%% pill
\begin{table*}
  \centering
  \begin{tabular}{lrr|rr|rr|rr}
    \hline
    Pill & \multicolumn{2}{c|}{Color} & \multicolumn{2}{c|}{Combined} & \multicolumn{2}{c|}{Contamination} & \multicolumn{2}{c}{Crack} \\
    Method & KID$\downarrow$ & LPIPS$\uparrow$ & KID$\downarrow$ & LPIPS$\uparrow$ & KID$\downarrow$ & LPIPS$\uparrow$ & KID$\downarrow$ & LPIPS$\uparrow$ \\
    \hline
    Finetune & 26.55 & 0.0762 & \textbf{14.56} & 0.0838 & \textbf{26.78} & 0.0751 & 55.56 & 0.0403 \\
    DiffAug   & 84.13 & 0.0525 & 98.90 & 0.0582 & 76.05 & 0.0360 & 76.39 & 0.0549 \\
    CDC   & 162.53 & 0.0508 & 120.11 & 0.0776 & 153.73 & 0.0562 & 147.60 & 0.0542 \\
    \hline
    Crop\&Paste   & - & \textbf{0.1201} & - & 0.1227 & - & 0.1226 & - & 0.1192 \\
    \hline
    SDGAN   & 106.42 & 0.0895 & 163.19 & 0.0796 & 75.74 & 0.0360 & 150.67 & 0.0775 \\
    Defect-GAN   & \textbf{2.64} & 0.0813 & 15.66 & 0.0890 & 31.07 & 0.0983 & \textbf{28.32} & 0.1042 \\
    \textbf{\name{}} & 126.95 & 0.1166 & 124.94 & \textbf{0.1441} & 99.72 & \textbf{0.1430} & 144.13 & \textbf{0.1573} \\
    \hline
  \end{tabular}
  \begin{tabular}{lrr|rr|rr}
    \hline
    pill & \multicolumn{2}{c|}{Faulty Imprint} & \multicolumn{2}{c|}{Pill Type} & \multicolumn{2}{c}{Scratch}\\
    Method & KID$\downarrow$ & LPIPS$\uparrow$ & KID$\downarrow$ & LPIPS$\uparrow$ & KID$\downarrow$ & LPIPS$\uparrow$ \\
    \hline
    Finetune & 28.10 & 0.0687 & \textbf{13.45} & 0.0743 & 23.30 & 0.0841 \\
    DiffAug   & 116.72 & 0.0396 & 30.04 & 0.0373 & 100.15 & 0.0431 \\
    CDC   & 259.56 & 0.0559 & 93.19 & 0.0681 & 114.36 & 0.0664 \\
    \hline
    Crop\&Paste   & - & 0.1091 & - & 0.0237 & -  & 0.1160\\
    \hline
    SDGAN   & 148.35 & 0.0884 & 279.53 & 0.0562 & 96.49 & 0.0658 \\
    Defect-GAN   & \textbf{14.85} & 0.1033 & 154.25 & 0.1211 & \textbf{15.94} & 0.0900 \\
    \textbf{\name{}} & 92.27 & \textbf{0.1523} & 141.80 & \textbf{0.2442} & 136.26 & \textbf{0.1436} \\
    \hline
  \end{tabular}
    \caption{The results of the few-shot defect image generation experiments on object category \emph{pill} with seven defect categories \emph{color}, \emph{combined}, \emph{contamination}, \emph{crack}, \emph{faulty imprint}, \emph{pill type}, and \emph{scratch}.}
    \label{tab:pill}
\end{table*}

\begin{table*}[t]
  \centering
  \begin{tabular}{lrrr}
    \hline
    pill & P1 Acc$\uparrow$ & P2 Acc$\uparrow$ & P3 Acc$\uparrow$ \\
    \hline
    Finetune & 23.96 & 19.79 & \textbf{33.33} \\
    DiffAug   & 30.21 & \textbf{34.38} & 25.00 \\
    CDC   & 22.92 & 20.83 & 21.88 \\
    \hline
    Crop\&Paste   & 28.13 & 26.04 & 26.04 \\
    \hline
    SDGAN   & 18.75 & 17.71 & 25.00 \\
    Defect-GAN   & \textbf{34.38} & 26.04 & 25.00 \\
    \textbf{\name{}} & 30.21 & 28.13 & 30.21 \\
    \hline
  \end{tabular}
    \caption{The results of the defect classification experiments on object category \emph{pill}.}
    \label{tab:pill-class}
\end{table*}

\clearpage

% double * 4
\begin{figure*}[t]
  \centering
  \includegraphics[width=\doublefourwidth\linewidth]{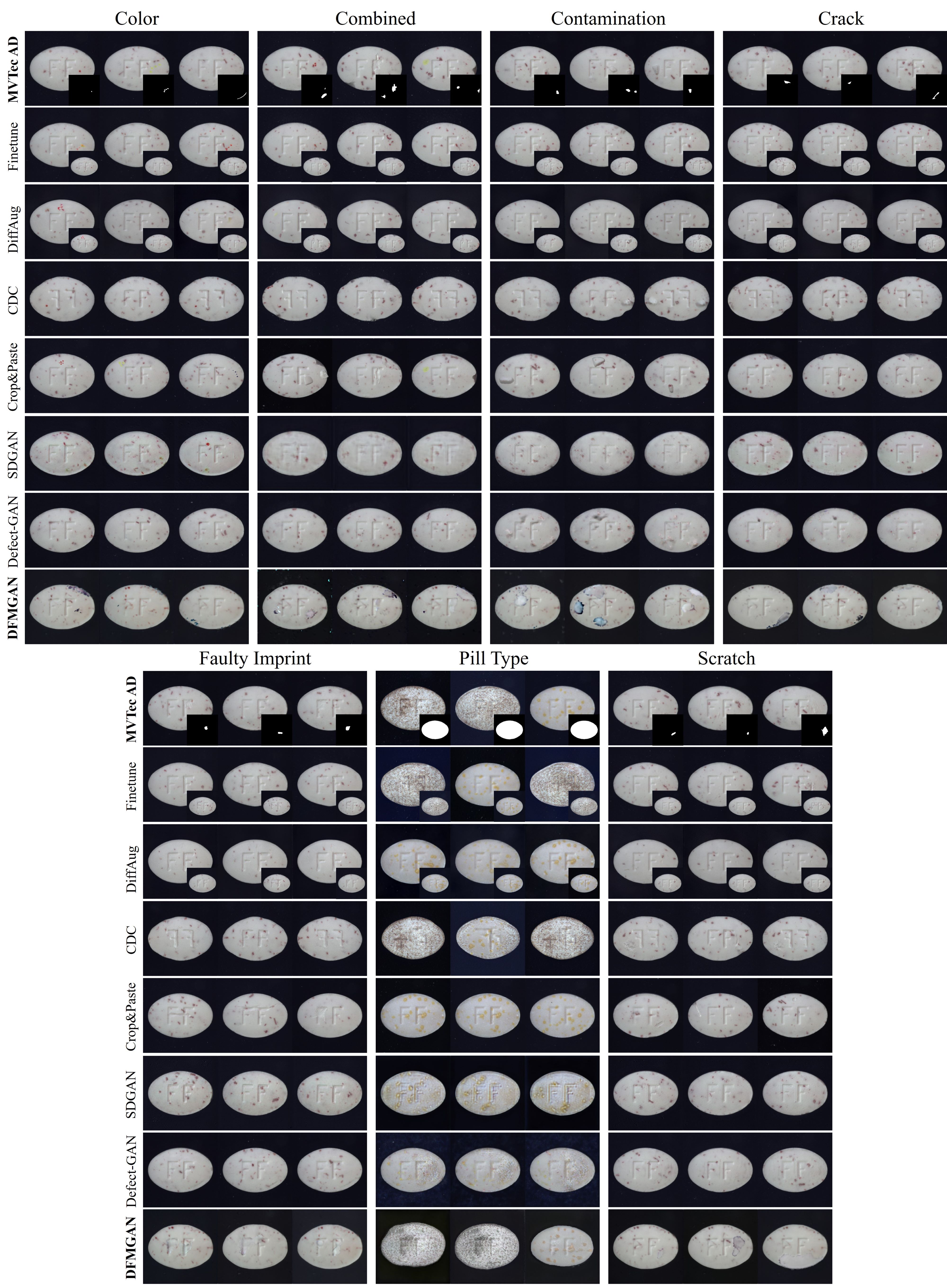}
   \caption{Examples of datasets (with masks) and generated defect images by different methods on object category \emph{pill}. }
   \label{fig:pill}
\end{figure*}

\clearpage

%%%%% screw
\begin{table*}
  \centering
  \begin{tabular}{lrr|rr|rr|rr|rr}
    \hline
    Screw & \multicolumn{2}{c|}{Manipulated Front} & \multicolumn{2}{c|}{Scratch Head} & \multicolumn{2}{c|}{Scratch Neck} & \multicolumn{2}{c|}{Thread Side} & \multicolumn{2}{c}{Thread Top} \\
    Method & KID$\downarrow$ & LPIPS$\uparrow$ & KID$\downarrow$ & LPIPS$\uparrow$ & KID$\downarrow$ & LPIPS$\uparrow$ & KID$\downarrow$ & LPIPS$\uparrow$ & KID$\downarrow$ & LPIPS$\uparrow$ \\
    \hline
    Finetune & 8.02 & 0.0901 & 5.64 & 0.1015 & 9.70 & 0.0892 & 9.45 & 0.1008 & 10.44 & 0.0887 \\
    DiffAug   & 12.31 & 0.1059 & 5.16 & 0.0894 & 37.74 & 0.1132 & 3.39 & 0.0948 & \textbf{2.33} & 0.0964 \\
    CDC   & 29.93 & 0.0992 & 28.21 & 0.1218 & 23.49 & 0.1162 & 24.26 & 0.1055 & 42.18 & 0.1125 \\
    \hline
    Crop\&Paste   & - & \textbf{0.1678} & - & \textbf{0.1668} & - & 0.1385 & - & \textbf{0.1689} & - & \textbf{0.1737} \\
    \hline
    SDGAN   & 96.76 & 0.1002 & 34.34 & 0.1026 & 94.73 & 0.0868 & 64.83 & 0.0919 & 120.69 & 0.0978 \\
    Defect-GAN   & 15.85 & 0.1155 & \textbf{2.24} & 0.1072 & \textbf{7.37} & 0.1223 & \textbf{1.20} & 0.1176 & 2.76 & 0.1339 \\
    \textbf{\name{}} & \textbf{7.93} & 0.1114 & 7.84 & 0.1464 & 11.51 & \textbf{0.1693} & 11.04 & 0.1424 & 9.35 & 0.1324 \\
    \hline
  \end{tabular}
    \caption{The results of the few-shot defect image generation experiments on object category \emph{screw} with five defect categories \emph{manipulated front}, \emph{scratch head}, \emph{scratch neck}, \emph{thread side} and \emph{thread top}.}
      \label{tab:screw}
\end{table*}

\begin{table*}[t]
  \centering
  \begin{tabular}{lrrr}
    \hline
    Screw & P1 Acc$\uparrow$ & P2 Acc$\uparrow$ & P3 Acc$\uparrow$ \\
    \hline
    Finetune & 29.63 & 27.16 & 23.46 \\
    DiffAug   & 27.16 & 24.69 & 23.46 \\
    CDC   & 33.33 & 35.80 & 29.63 \\
    \hline
    Crop\&Paste   & 30.86 & 29.63 & 25.93 \\
    \hline
    SDGAN   & 28.40 & 32.09 & 19.75 \\
    Defect-GAN   & 34.57 & 22.22 & 29.63 \\
    \textbf{\name{}} & \textbf{35.80} & \textbf{43.21} & \textbf{33.33}  \\
    \hline
  \end{tabular}
    \caption{The results of the defect classification experiments on object category \emph{screw}.}
    \label{tab:screw-class}
\end{table*}

\clearpage

% vertical * 5
\begin{figure*}[t]
  \centering
  \includegraphics[angle=-90, width=\fivewidth \linewidth]{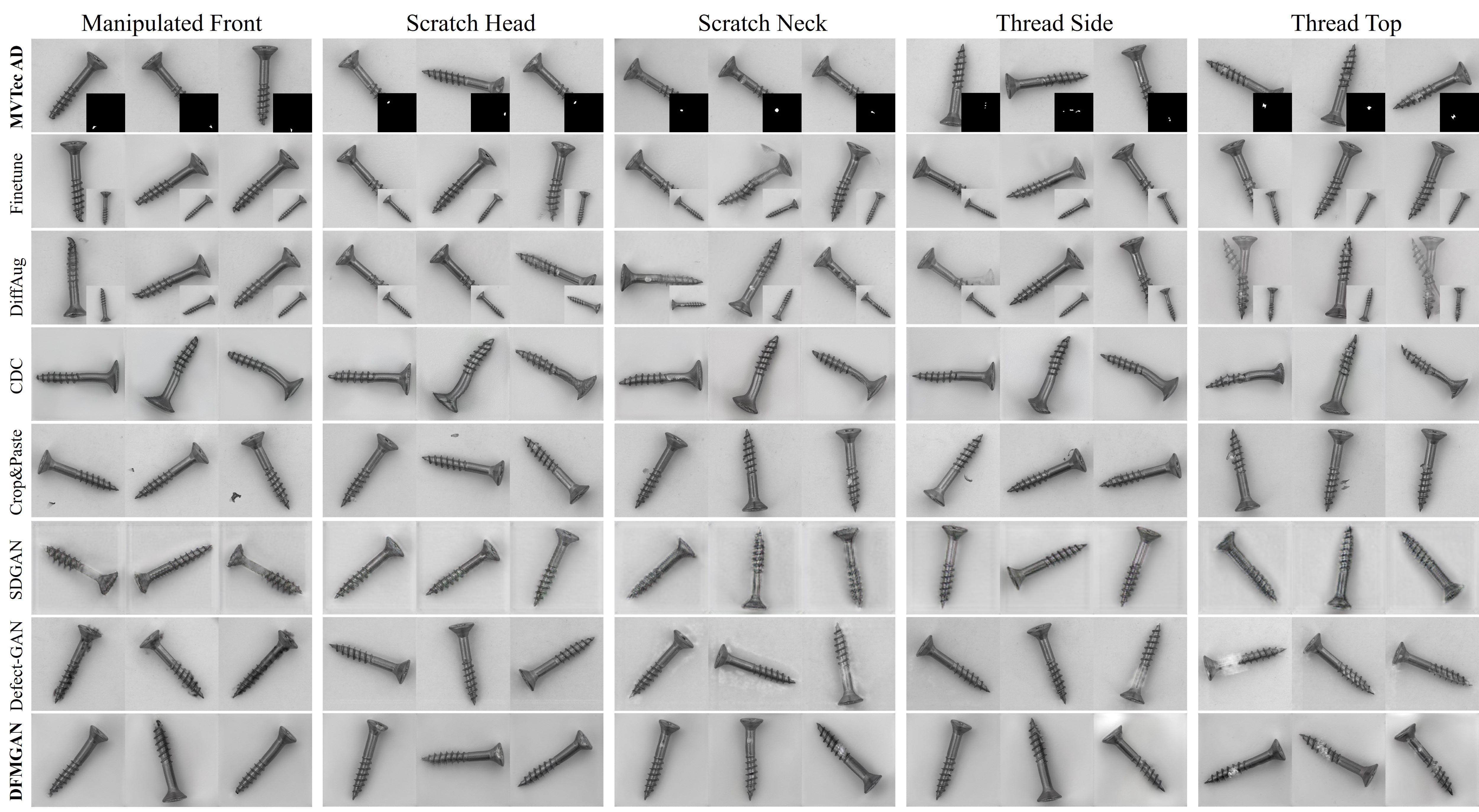}
   \caption{Examples of datasets (with masks) and generated defect images by different methods on object category \emph{screw}. }
   \label{fig:screw}
\end{figure*}

\clearpage

%%%%% tile
\begin{table*}
  \centering
  \begin{tabular}{lrr|rr|rr|rr|rr}
    \hline
    Tile & \multicolumn{2}{c|}{Crack} & \multicolumn{2}{c|}{Glue Strip} & \multicolumn{2}{c|}{Gray Stroke} & \multicolumn{2}{c|}{Oil} & \multicolumn{2}{c}{Rough} \\
    Method & KID$\downarrow$ & LPIPS$\uparrow$ & KID$\downarrow$ & LPIPS$\uparrow$ & KID$\downarrow$ & LPIPS$\uparrow$ & KID$\downarrow$ & LPIPS$\uparrow$ & KID$\downarrow$ & LPIPS$\uparrow$ \\
    \hline
    Finetune & \textbf{62.99} & 0.1225 & 123.66 & 0.1347 & \textbf{34.71} & 0.1025 & \textbf{51.95} & 0.1116 & \textbf{47.51} & 0.1305 \\
    DiffAug   & 130.87 & 0.0835 & 296.25 & 0.1211 & 149.94 & 0.0571 & 181.01 & 0.0659 & 171.66 & 0.1173 \\
    CDC   & 179.75 & 0.0912 & 203.31 & 0.1118 & 41.53 & 0.1282 & 93.75 & 0.1060 & 49.57 & 0.1418 \\
    \hline
    Crop\&Paste   & - & 0.1872 & - & 0.2028 & - & \textbf{0.2204} & - & 0.1845 & - & 0.1892 \\
    \hline
    SDGAN   & 511.18 & \textbf{0.2703} & 412.33 & 0.1918 & 364.13 & 0.1180 & 340.97 & 0.2428 & 381.82 & 0.2180 \\
    Defect-GAN   & 452.90 & 0.2254 & 200.57 & 0.2199 & 48.81 & 0.1827 & 245.56 & \textbf{0.2458} & 98.16 & 0.2409 \\
    \textbf{\name{}} & 131.27 & 0.2175 & \textbf{105.16} & \textbf{0.2361} & 46.34 & 0.2024 & 68.27 & 0.2094 & 75.34 & \textbf{0.2417} \\
    \hline
  \end{tabular}
    \caption{The results of the few-shot defect image generation experiments on texture category \emph{tile} with five defect categories \emph{crack}, \emph{glue strip}, \emph{gray stroke}, \emph{oil} and \emph{rough}.}
    \label{tab:tile}
\end{table*}

\begin{table*}[t]
  \centering
  \begin{tabular}{lrrr}
    \hline
    Tile & P1 Acc$\uparrow$ & P2 Acc$\uparrow$ & P3 Acc$\uparrow$ \\
    \hline
    Finetune & 59.65 & 52.63 & 45.61 \\
    DiffAug   & 59.65 & 56.14 & 63.16 \\
    CDC   & 33.33 & 59.65 & 52.63 \\
    \hline
    Crop\&Paste   & 68.42 & 71.93 & 64.91 \\
    \hline
    SDGAN   & 45.61 & 47.37 & 35.09 \\
    Defect-GAN   & 24.56 & 24.56 & 31.58 \\
    \textbf{\name{}} & \textbf{75.44} & \textbf{75.44} & \textbf{73.68} \\
    \hline
  \end{tabular}
    \caption{The results of the defect classification experiments on texture category \emph{tile}.}
    \label{tab:tile-class}
\end{table*}

\clearpage

% vertical * 5
\begin{figure*}[t]
  \centering
  \includegraphics[angle=-90, width=\fivewidth \linewidth]{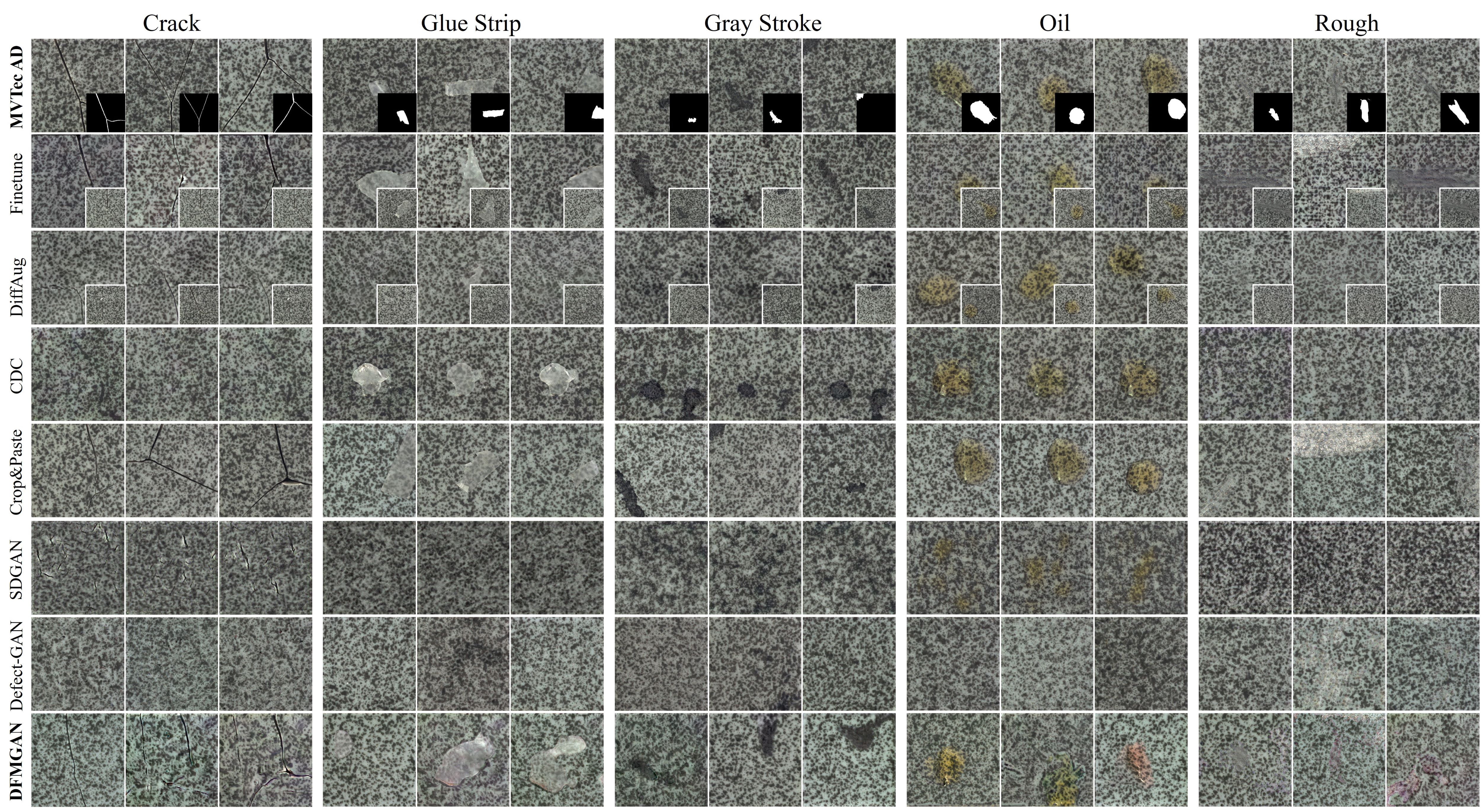}
   \caption{Examples of datasets (with masks) and generated defect images by different methods on texture category \emph{tile}.   }
   \label{fig:tile}
\end{figure*}

\clearpage

%%%%% toothbrush
\begin{table*}[t]
  \centering
  \begin{tabular}{lrr}
    \hline
    toothbrush & \multicolumn{2}{c}{Defective} \\
    Method & KID$\downarrow$ & LPIPS$\uparrow$ \\
    \hline
    Finetune & 39.40 & 0.0461 \\
    DiffAug   & \textbf{13.89} & 0.0632 \\
    CDC   & 22.87 & 0.0623 \\
    \hline
    Crop\&Paste   & - & 0.0757 \\
    \hline
    SDGAN   & 187.88 & 0.0297 \\
    Defect-GAN   & 37.20 & 0.0306 \\
    \textbf{\name{}} & 46.49 & \textbf{0.1839} \\
    \hline
  \end{tabular}
    \caption{The results of the few-shot defect image generation experiments on object category \emph{toothbrush} with one defect category \emph{defective}.}
    \label{tab:toothbrush}
\end{table*}

% vertical * 5
\begin{figure*}[t]
  \centering
  \includegraphics[angle=-90, width=\fivewidth \linewidth]{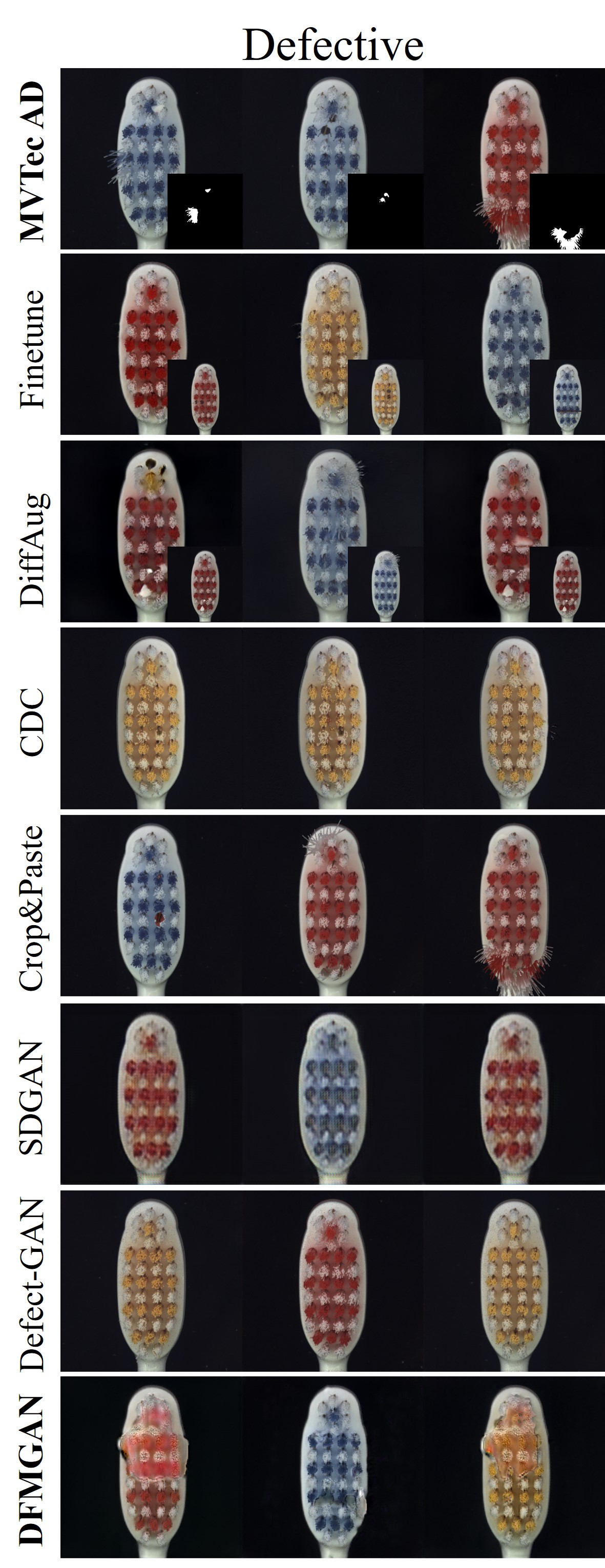}
   \caption{Examples of datasets (with masks) and generated defect images by different methods on object category \emph{toothbrush}. }
   \label{fig:toothbrush}
\end{figure*}

\clearpage

%%%%% transistor
\begin{table*}[t]
  \centering
  \begin{tabular}{lrr|rr|rr|rr}
    \hline
    Transistor & \multicolumn{2}{c|}{Bent Lead} & \multicolumn{2}{c|}{Cut Lead} & \multicolumn{2}{c|}{Damaged Case} & \multicolumn{2}{c}{Misplaced} \\
    Method & KID$\downarrow$ & LPIPS$\uparrow$ & KID$\downarrow$ & LPIPS$\uparrow$ & KID$\downarrow$ & LPIPS$\uparrow$ & KID$\downarrow$ & LPIPS$\uparrow$ \\
    \hline
    Finetune & 43.38 & 0.1340 & 37.65 & 0.1135 & 33.52 & 0.0781 & 19.98 & 0.1282 \\
    DiffAug   & 24.19 & 0.0633 & 18.26 & 0.0435 & \textbf{7.67} & 0.0359 & \textbf{3.47} & 0.0625 \\
    CDC   & \textbf{16.94} & 0.1343 & \textbf{8.64} & 0.1191 & 19.75 & 0.1189 & 106.95 & 0.1484 \\
    \hline
    Crop\&Paste   & - & 0.1879 & - & 0.1465 & - & 0.1784 & - & 0.0873 \\
    \hline
    SDGAN   & 171.43 & 0.1439 & 194.75 & 0.1394 & 151.43 & 0.0999 & 241.06 & 0.1266 \\
    Defect-GAN   & 71.37 & 0.1491 & 108.40 & 0.0435 & 107.13 & 0.1405 & 159.16 & 0.1881 \\
    \textbf{\name{}} & 62.37 & \textbf{0.2351} & 82.63 & \textbf{0.2232} & 98.63 & \textbf{0.2408} & 109.59 & \textbf{0.2936} \\
    \hline
  \end{tabular}
    \caption{The results of the few-shot defect image generation experiments on object category \emph{transistor} with four defect categories \emph{bent lead}, \emph{cut lead}, \emph{damaged case} and \emph{misplaced}.}
     \label{tab:transistor}
\end{table*}

\begin{table*}[t]
  \centering
  \begin{tabular}{lrrr}
    \hline
    Transistor & P1 Acc$\uparrow$ & P2 Acc$\uparrow$ & P3 Acc$\uparrow$ \\
    \hline
    Finetune & 39.29 & 32.14 & 28.57 \\
    DiffAug   & 42.86 & 35.71 & 35.71 \\
    CDC   & 25.00 & 28.57 & 35.71 \\
    \hline
    Crop\&Paste   & 42.86 & 39.29 & 42.86 \\
    \hline
    SDGAN   & 35.71 & 25.00 & 35.71 \\
    Defect-GAN   & 28.57 & 39.29 & 39.29 \\
    \textbf{\name{}} & \textbf{57.14} & \textbf{42.86} & \textbf{57.14} \\
    \hline
  \end{tabular}
    \caption{The results of the defect classification experiments on object category \emph{transistor}.}
    \label{tab:transistor-class}
\end{table*}

\begin{figure*}[t]
  \centering
  \includegraphics[width=\singlefourwidth\linewidth]{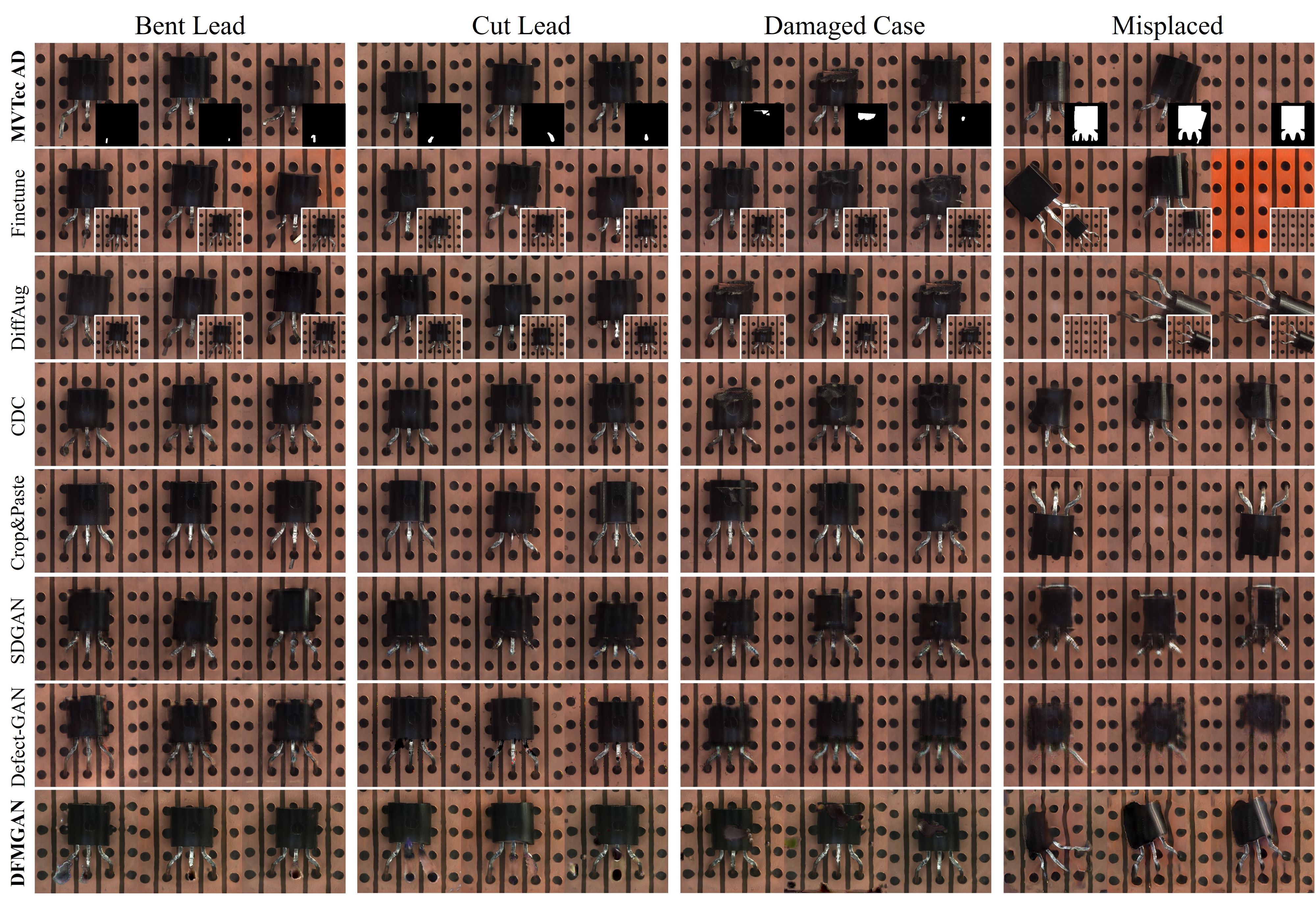}
   \caption{Examples of datasets (with masks) and generated defect images by different methods on object category \emph{transistor}. }
   \label{fig:transistor}
\end{figure*}

\clearpage

%%%%% wood
\begin{table*}
  \centering
  \begin{tabular}{lrr|rr|rr|rr|rr}
    \hline
    Wood & \multicolumn{2}{c|}{Color} & \multicolumn{2}{c|}{Combined} & \multicolumn{2}{c|}{Hole} & \multicolumn{2}{c|}{Liquid} & \multicolumn{2}{c}{Scratch} \\
    Method & KID$\downarrow$ & LPIPS$\uparrow$ & KID$\downarrow$ & LPIPS$\uparrow$ & KID$\downarrow$ & LPIPS$\uparrow$ & KID$\downarrow$ & LPIPS$\uparrow$ & KID$\downarrow$ & LPIPS$\uparrow$ \\
    \hline
    Finetune & 56.27 & 0.2895 & 29.10 & 0.3321 & 24.68 & 0.3131 & 38.42 & 0.2821 & 26.37 & 0.3429 \\
    DiffAug   & \textbf{24.92} & 0.2683 & \textbf{28.29} & 0.3417 & \textbf{15.10} & 0.3087 & \textbf{15.14} & 0.2639 & \textbf{19.09} & 0.3379 \\
    CDC   & 141.67 & 0.0183 & 83.14 & 0.0289 & 183.96 & 0.0247 & 218.01 & 0.0328 & 133.60 & 0.0545 \\
    \hline
    Crop\&Paste   & - & 0.2383 & - & 0.2275 & - & 0.2196 & - & 0.2456 & - & 0.2256 \\
    \hline
    SDGAN   & 364.04 & 0.2458 & 365.27 & 0.2450 & 178.54 & 0.2836 & 286.74 & 0.2094 & 197.75 & 0.2659  \\
    Defect-GAN   & 135.23 & 0.2732 & 150.83 & 0.3172 & 100.61 & 0.2866 & 190.29 & 0.2776 & 67.35 & 0.2946 \\
    \textbf{\name{}} & 74.30 & \textbf{0.3649} & 68.09 & \textbf{0.3617} & 78.07 & \textbf{0.3243} & 59.84 & \textbf{0.3402} & 60.32 & \textbf{0.3432} \\
    \hline
  \end{tabular}
    \caption{The results of the few-shot defect image generation experiments on texture category \emph{wood} with five defect categories \emph{color}, \emph{combined}, \emph{hole}, \emph{liquid} and \emph{scratch}.}
    \label{tab:wood}
\end{table*}

\begin{table*}[t]
  \centering
  \begin{tabular}{lrrr}
    \hline
    Wood & P1 Acc$\uparrow$ & P2 Acc$\uparrow$ & P3 Acc$\uparrow$ \\
    \hline
    Finetune & 38.10 & 40.48 & 42.86 \\
    DiffAug   & 40.48 & 40.48 & 42.86 \\
    CDC   & 33.33 & 28.57 & 23.81 \\
    \hline
    Crop\&Paste   & 45.23 & \textbf{47.62} & 50.00\\
    \hline
    SDGAN   & 28.57 & 38.10 & 26.19 \\
    Defect-GAN   & 28.57 & 19.05 & 26.19 \\
    \textbf{\name{}} & \textbf{47.62} & 45.24 & \textbf{54.76} \\
    \hline
  \end{tabular}
    \caption{The results of the defect classification experiments on texture category \emph{wood}.}
    \label{tab:wood-class}
\end{table*}

\clearpage

\begin{figure*}[t]
  \centering
  \includegraphics[angle=-90, width=\fivewidth \linewidth]{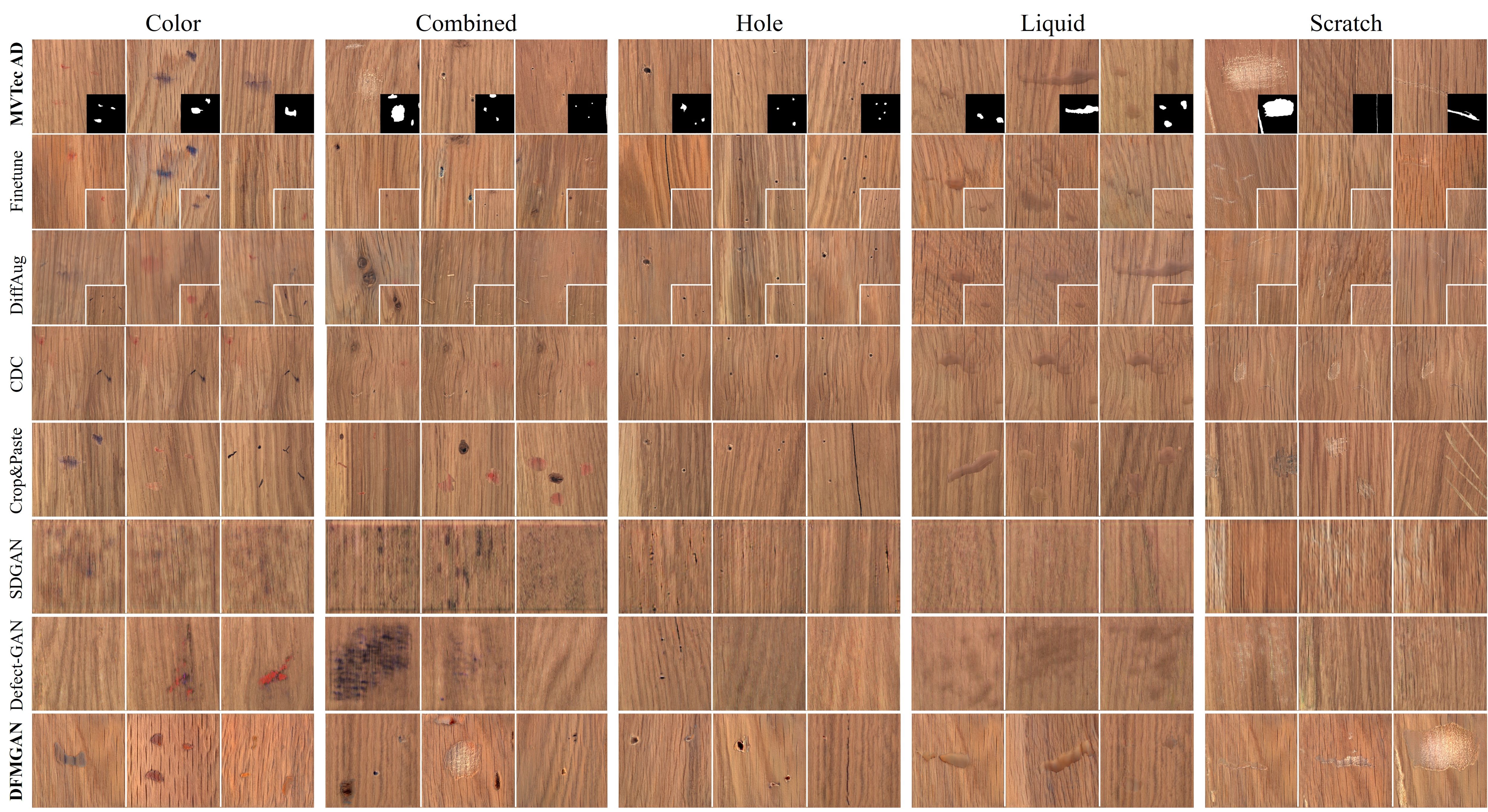}
   \caption{Examples of datasets (with masks) and generated defect images by different methods on texture category \emph{wood}.   }
   \label{fig:wood}
\end{figure*}

\clearpage

%%%%% zipper
\begin{table*}
  \centering
  \begin{tabular}{lrr|rr|rr|rr}
    \hline
    Zipper & \multicolumn{2}{c|}{Broken Teeth} & \multicolumn{2}{c|}{Combined} & \multicolumn{2}{c|}{Fabric Border} & \multicolumn{2}{c}{Fabric Interior} \\
    Method & KID$\downarrow$ & LPIPS$\uparrow$ & KID$\downarrow$ & LPIPS$\uparrow$ & KID$\downarrow$ & LPIPS$\uparrow$ & KID$\downarrow$ & LPIPS$\uparrow$ \\
    \hline
    Finetune & 74.91 & 0.0125 & 110.45 & 0.0118 & 43.96 & 0.0282 & 43.05 & 0.0146 \\
    DiffAug   & \textbf{37.53} & 0.0166 & \textbf{23.03} & 0.0991 & \textbf{22.98} & 0.0598 & \textbf{28.33} & 0.0258 \\
    CDC   & 286.99 & 0.0298 & 106.11 & 0.0411 & 70.43 & 0.0519 & 141.88 & 0.0403 \\
    \hline
    Crop\&Paste   & - & 0.1071 & - & 0.1142 & - & 0.1247 & - & 0.0893 \\
    \hline
    SDGAN   & 96.91 & 0.0963 & 85.94 & 0.0992 & 100.59 & 0.0974 & 81.03 & 0.0974 \\
    Defect-GAN   & 59.66 & 0.0918 & 87.98 & 0.0910 & 38.66 & 0.0774 & 73.57 & 0.1073 \\
    \textbf{\name{}} & 76.00 & \textbf{0.3049} & 75.20 & \textbf{0.2605} & 61.83 & \textbf{0.2894} & 63.21 & \textbf{0.2499} \\
    \hline
  \end{tabular}
  \begin{tabular}{lrr|rr|rr}
    \hline
    Zipper & \multicolumn{2}{c|}{Rough} & \multicolumn{2}{c|}{Split Teeth} & \multicolumn{2}{c}{Squeezed Teeth}\\
    Method & KID$\downarrow$ & LPIPS$\uparrow$ & KID$\downarrow$ & LPIPS$\uparrow$ & KID$\downarrow$ & LPIPS$\uparrow$ \\
    \hline
    Finetune & 85.36 & 0.0122 & 44.38 & 0.0280 & 85.57 & 0.0134 \\
    DiffAug   & \textbf{29.90} & 0.0309 & \textbf{11.05} & 0.0415 & \textbf{7.70} & 0.0473 \\
    CDC   & 41.07 & 0.0332 & 55.05 & 0.0326 & 119.79 & 0.0959 \\
    \hline
    Crop\&Paste   & - & 0.1162 & - & 0.1025 & - & 0.1021 \\
    \hline
    SDGAN   & 96.36 & 0.1052 & 140.67 & 0.0977 & 119.79 & 0.0959 \\
    Defect-GAN   & 41.26 & 0.0979 & 55.50 & 0.1245 & 139.41 & 0.1324 \\
    \textbf{\name{}} & 43.30 & \textbf{0.2604} & 119.49 & \textbf{0.2122} & 104.66 & \textbf{0.2864} \\
    \hline
  \end{tabular}
    \caption{The results of the few-shot defect image generation experiments on object category \emph{zipper} with seven defect categories \emph{broken teeth}, \emph{combined}, \emph{fabric border}, \emph{fabric interior}, \emph{rough}, \emph{split teeth}, and \emph{squeezed teeth}.}
    \label{tab:zipper}
\end{table*}

\begin{table*}[t]
  \centering
  \begin{tabular}{lrrr}
    \hline
    zipper & P1 Acc$\uparrow$ & P2 Acc$\uparrow$ & P3 Acc$\uparrow$ \\
    \hline
    Finetune & 18.29 & 20.73 & 14.63 \\
    DiffAug   & 23.17 & 24.39 & 20.73 \\
    CDC   & 12.19 & 18.29 & 13.41 \\
    \hline
    Crop\&Paste   & \textbf{30.49} & 26.83 & 21.95 \\
    \hline
    SDGAN   & 23.17 & 23.17 & 18.29 \\
    Defect-GAN   & 25.61 & 13.41 & 17.07 \\
    \textbf{\name{}} & \textbf{30.49} & \textbf{28.05} & \textbf{24.39} \\
    \hline
  \end{tabular}
    \caption{The results of the defect classification experiments on object category \emph{zipper}.}
    \label{tab:zipper-class}
\end{table*}

\clearpage

% double * 4
\begin{figure*}[t]
  \centering
  \includegraphics[width=\doublefourwidth\linewidth]{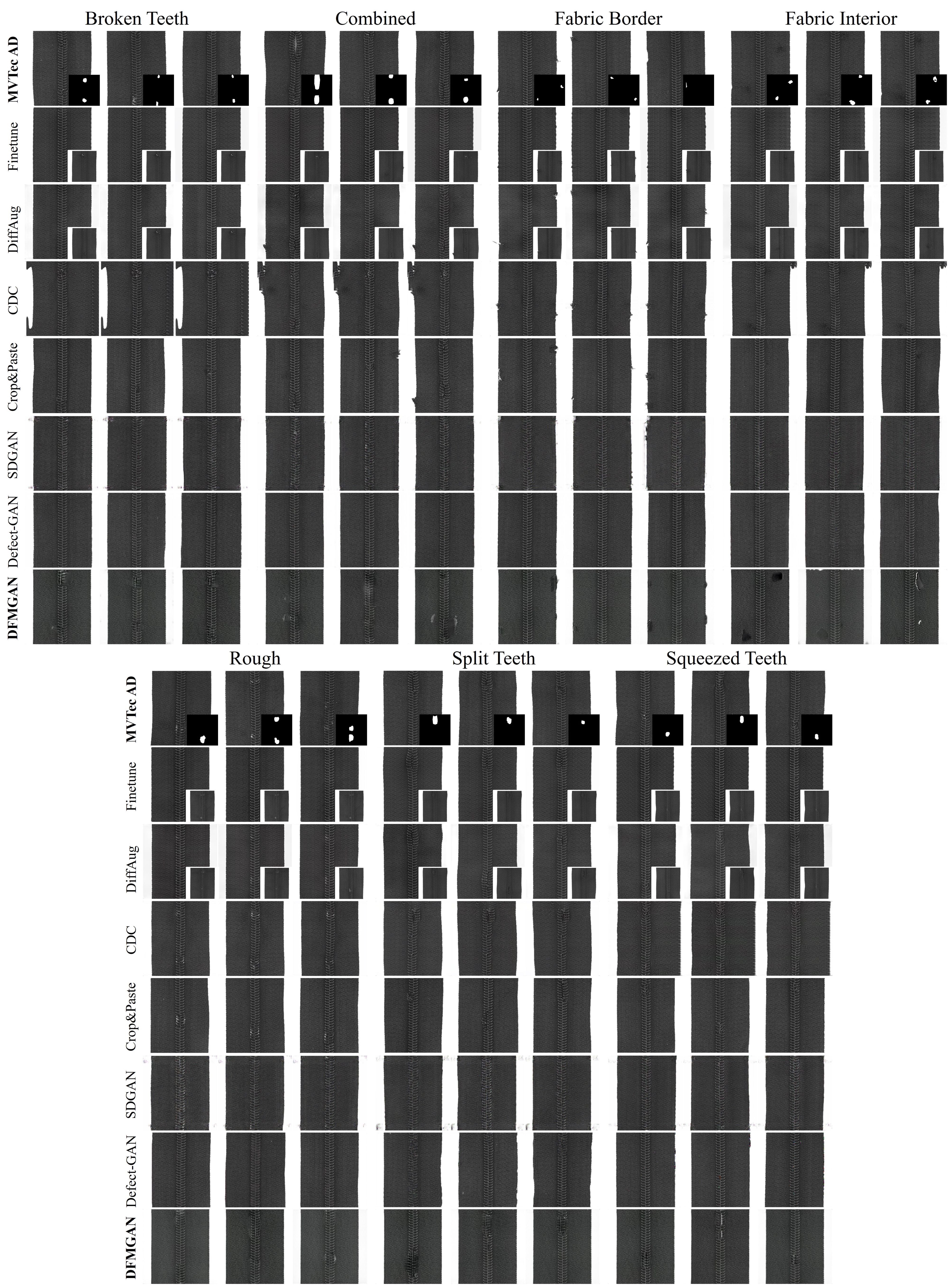}
   \caption{Examples of datasets (with masks) and generated defect images by different methods on object category \emph{zipper}.   }
   \label{fig:zipper}
\end{figure*}

%%%%% COMMENT
% \begin{figure*}[t]
%   \centering
%   \includegraphics[width=0.9\linewidth]{figures/hazelnut_itp_3k.png}
%   \caption{Examples of generated defect images on object category \emph{hazelnut}. We fix/interpolate the two random codes $\zobject$ and $\zdefect$.}
%   \label{fig:hazelnut-itp}
% \end{figure*}

% \begin{figure*}[t]
%   \centering
%   \includegraphics[width=0.9\linewidth]{figures/metalnut_itp_3k.png}
%   \caption{Examples of generated defect images on object category \emph{metalnut}. We fix/interpolate the two random codes $\zobject$ and $\zdefect$.}
%   \label{fig:metalnut-itp}
% \end{figure*}

% \begin{figure*}[t]
%   \centering
%   \includegraphics[width=0.9\linewidth]{figures/wood_itp_3k.png}
%   \caption{Examples of generated defect images on texture category \emph{wood} with fixed/interpolated random codes $\zobject$ and $\zdefect$.  }
%   \label{fig:wood-itp}
% \end{figure*}
%%%%% COMMENT

% Use \bibliography{yourbibfile} instead or the References section will not appear in your paper